\documentclass[twoside,11pt]{article}

\usepackage{paralist,amsmath, amssymb}
\usepackage{booktabs}
\usepackage{multirow}
\usepackage{algorithm,algorithmic}
\usepackage{jmlr2e}

\usepackage[colorlinks,
            linkcolor=blue,
            citecolor=blue,
            urlcolor=magenta,
            linktocpage,
            plainpages=false]{hyperref}

\makeatletter
\newcounter{ALC@tempcntr}%
\makeatother

\makeatletter
\AtBeginDocument{\Hy@breaklinkstrue}
\makeatother

\usepackage{mathtools}
\usepackage{bm}
\usepackage{dsfont}
\usepackage{amsfonts}
\def \EndProof {\hfill $\square$\par}
\def \BeginProof {{\noindent\it\bfseries Proof.\quad}}
\def \y {\mathbf{y}}
\def \E {\mathrm{E}}
\def \EP {\mathcal{E}}
\def \x {\mathbf{x}}

\def \D {\mathcal{D}}
\def \O {\mathcal{O}}
\def \z {\mathbf{z}}
\def \u {\mathbf{u}}

\def \w {\mathbf{w}}

\def \A {\mathcal{A}}

\def \v {\mathbf{v}}

\def \p {\mathbf{p}}

\def \I {\mathcal{I}}

\def \a {\mathbf{a}}
\def \b {\mathbf{b}}

\def \B {\mathbf{B}}

\def \C {\mathcal{C}}
\def \U {\mathcal{U}}

\def \lh {\bm{\ell}}
\def \e {\mathbf{e}}
\def \B {\mathcal{B}}
\def \E {\mathcal{E}}
\def \X {\mathcal{X}}

\DeclareMathOperator*{\SR}{SR}

\DeclareMathOperator*{\TR}{TR}
\DeclareMathOperator*{\tr}{Tr}
\DeclareMathOperator*{\diag}{diag}

\newtheorem{thm}{Theorem}
\newtheorem{prop}{Proposition}
\newtheorem{lma}{Lemma}

\newtheorem{defi}{Definition}
\newtheorem{rmk}{Remark}

\begin{document}

\title{Adaptive and Efficient Algorithms for \\ Tracking the Best Expert}

\author{Shiyin Lu \email lusy@lamda.nju.edu.cn\\
\name Lijun Zhang \email zhanglj@lamda.nju.edu.cn\\
       \addr National Key Laboratory for Novel Software Technology\\
       Nanjing University, Nanjing 210023, China
}

\editor{}

\maketitle

\begin{abstract}%
In this paper, we consider the problem of prediction with expert advice in dynamic environments. We choose \emph{tracking} regret as the performance metric and develop two adaptive and efficient algorithms with data-dependent tracking regret bounds. The first algorithm achieves a second-order tracking regret bound, which improves existing first-order bounds. The second algorithm enjoys a path-length bound, which is generally not comparable to the second-order bound but offers advantages in slowly moving environments. Both algorithms are developed under the online mirror descent framework and draw inspiration from existing algorithms that attain data-dependent bounds of \emph{static} regret. The key idea is to use a clipped simplex in the updating step of online mirror descent. Finally, we extend our algorithms and analysis to online matrix prediction and provide the \emph{first} data-dependent tracking regret bound for this problem.
\end{abstract}


\section{Introduction}
We study the problem of prediction with expert advice, where a learner makes sequential predictions by combining advice from $K$ experts. We consider the following decision-theoretic setup \citep{FREUND1997119}: In each round $t=1,\ldots,T$, the learner chooses a distribution $\w_t$ over $K$ experts, and at the same time an adversary decides a loss vector $\lh_t$ encoding the loss of each expert: $\lh_t = (\lh_t[1], \ldots, \lh_t[K]) \in [0, 1]^K$. Then, the learner observes the loss vector $\lh_t$ and suffers a weighted average loss $\hat \ell_t = \langle \w_t, \lh_t \rangle$, where $\langle \cdot, \cdot \rangle$ denotes the inner product. The classic metric to measure the learner's performance is \emph{static} regret, defined as the difference between the cumulative loss of the learner and that of the best single expert over $T$ rounds in hindsight: 
\begin{equation*}
   \SR(T) = \sum_{t=1}^T \hat \ell_t - \min_{\EP \in [K] } \sum_{t=1}^T \lh_t[\EP]
\end{equation*}
where $[K] = \{1, 2, \ldots, K\}$.
During the past decades, minimizing static regret has been extensively studied, and minimax-optimal algorithms with $\O(\sqrt{T\log{K}})$ regret bounds as well as adaptive algorithms with data-dependent regret bounds have been developed \citep{bianchi-2006-prediction}. However, the static regret is only meaningful for stationary environments where a single expert performs well over $T$ rounds, and fails to illustrate the performance of online algorithms in changing environments where the best expert could switch over time.

To address this limitation, a more stringent metric called \emph{tracking} regret has been introduced and studied in the literature under the name of ``tracking the best expert'' \citep{Herbster1998,Vovk1999,Herbster:2001:TBL,bousquet2002tracking}. Instead of competing with a single expert, in tracking regret the learner is compared against a sequence of experts $\EP_1, \ldots, \EP_T$ with a small number of switches $\EP_t \neq \EP_{t-1}$:
\begin{equation}
   \label{eq:def:tr}
   \begin{split}
      \TR(T, S)  = \sum_{t=1}^T \hat \ell_t - \min_{(\EP_1, \ldots, \EP_T) \in \C(T, S)} \sum_{t=1}^T \lh_t[\EP_t] 
       = \sum_{t=1}^T \hat \ell_t - \sum_{t=1}^T \lh_t[\EP_t^*]
   \end{split}
\end{equation}
where $\C(T, S)$ is the set comprised of all sequence of experts in which the expert switches at most $S - 1$ times:
\begin{equation}
   \label{def:CTS}
   \begin{split}
      \C(T, S) = \left\{  (\EP_1, \ldots, \EP_T) \in [K]^T ~\vert~ 
       \sum_{t=2}^T \mathds{1}{\{\EP_t \neq \EP_{t-1}\}} \leq S - 1\right\}
   \end{split}
\end{equation}
and $\EP_1^*,\ldots,\EP_T^*$ is the best sequence of experts in $\C(T, S)$:
\begin{equation}
   \label{def:best-sequence-of-experts}
   (\EP_1^*,\ldots,\EP_T^*) = \mathop{\arg\min}_{(\EP_1, \ldots, \EP_T) \in \C(T, S)} \sum_{t=1}^T \lh_t[\EP_t] .
\end{equation}
It is easy to see that the tracking regret includes the static regret as a special case by setting $S=1$.

As early as 20 years ago, \citet{Herbster1998,Herbster:2001:TBL} have developed two algorithms for tracking the best expert, namely, fixed share and projection update, both of which enjoy an $\O\big(\sqrt{ST\log{(KT/S)}}\big)$ tracking regret bound. While this bound is not improvable in general, we are interested in obtaining more favorable data-dependent bounds of \emph{tracking} regret, which match the above  bound in the worst case but become much smaller in benign environments. 
In this paper, we present two adaptive and efficient algorithms that enjoy data-dependent tracking regret bounds. The first algorithm is shown to achieve a novel \emph{second-order} tracking regret bound of $\O\big(\sqrt{SL_2\log{(KT/S)}} + S\log{(KT/S)} \big)$, where $L_2$ is the sum of squared loss of $\EP_1^*,\ldots,\EP_T^*$:
\begin{equation}
   \label{def:L2}
   L_2 = \sum_{t=1}^T (\lh_t[\EP_t^*])^2 .
\end{equation}
The second algorithm attains a \emph{path-length} bound of $\O\big(\sqrt{SP_{\infty}\log{(KT/S)}} + S\big)$, where $P_{\infty}$ is the sum of the square of the difference between consecutive loss vectors $\lh_1, \ldots, \lh_T$:
\begin{equation}
   \label{def:Pinfty}
   P_{\infty} = \sum_{t=1}^T \Vert \lh_t - \lh_{t-1} \Vert_{\infty}^2
\end{equation}
with $\lh_0 = (0, \ldots ,0)$. While this bound has been derived by previous work \citep{wei2016tracking}, their algorithm is inefficient since it needs to maintain $KT$ virtual experts and update the weights of $Kt$ experts in round $t$, which makes the space and time complexities per round grow linearly with $t$. By contrast, our algorithm performs on $K$ real experts and thus its space and time complexities per round are independent of $t$.

The second-order and the path-length bounds are not comparable in general and each has its own advantage: The former is better in the case that the loss of the best sequence of experts is small, while the latter exhibits superiority when the loss of all experts (i.e., the loss vector) moves slowly with time. Nevertheless, our second-order bound is better than the existing first-order bound \citep{Fixed:Share:NIPS12} of $\O\big(\sqrt{SL_1\log{(KT/S)}} + S\log{(KT/S)}\big)$ with $L_1 = \sum_{t=1}^T \lh_t[\EP_t^*]$, since the loss of experts is in the range of $[0, 1]$.

Both of our algorithms fall into the online mirror descent (OMD) framework \citep{Online:suvery} and are inspired by existing algorithms that enjoy data-dependent static regret bounds \citep{Cesa-Bianchi2005,pmlr-v23-chiang12}. The key technique is that in the updating step of OMD, we restrict the feasible set to be a clipped simplex to ensure the distribution assigned to each expert is lower bounded by a constant. While this technique can be shown as a different form of projection update, its advantage is that the intermediate distribution appearing in projection update is avoided and thus, we can analyze our algorithms under the framework of OMD. We also re-derive the Prod method \citep{Cesa-Bianchi2005}, which enjoys the second-order static regret bound, in the OMD framework so that the technique of clipped simplex can be applied. Finally, we present extensions of our algorithms and analysis to online matrix prediction and establish the \emph{first} data-dependent tracking regret bound for this problem.

\section{Related Work}
In this section, we briefly review related work on prediction with expert advice.
\subsection{Static Regret}
In their seminal work, \citet{LITTLESTONE1994212} and \citet{Vovk1990Aggregating} introduced the multiplicative weights update (MWU) method, also known as the exponentiated gradient (EG) algorithm \citep{kivinen1997exponentiated} and the Hedge algorithm \citep{FREUND1997119}. Starting from a uniform distribution $\w_1 = (1/K, 1/K, \ldots, 1/K)$, at each round $t$, MWU updates the distribution as
\begin{equation}
   \label{eq:mwu-update}
   \w_{t+1}[i] = \frac{\w_t[i] \exp(-\eta \lh_t[i])}{\sum_{j=1}^K  \w_t[j] \exp(-\eta \lh_t[j])}, \, \forall i \in [K]
\end{equation}
where $\eta$ is the learning rate. MWU was known to enjoy the first-order bound on static regret \citep{FREUND1997119}. Such bound is also attainable for the follow the perturbed leader (FPL) method \citep{hannan1957approximation,kalai2003efficient}, where the distribution is chosen based on the observed past loss vectors and a random generated loss vector. \citet{Cesa-Bianchi2005} proposed the Prod algorithm where the exponential update $\w_{t+1}[i] \propto \w_t[i] \exp(-\eta \lh_t[i])$ in MWU is replaced with the so-called multilinear update $\w_{t+1}[i] \propto \w_t[i] (1-\eta \lh_t[i])$, and showed that Prod achieves the second-order bound on static regret. While both the first-order and the second-order bounds belong to the family of small-loss bounds, there also exist other classes of data-dependent bounds. \citet{variation:bound} derived the variance bound which depends on the deviation of the loss vector from its average. \citet{pmlr-v23-chiang12} showed that a variant of MWU achieves the path-length bound. While the second-order bound is better than the first-order bound, except for the first-order bound, the other three bounds are not comparable in general \citep{steinhardt2014adaptivity}.

\subsection{Tracking Regret}
Two classic algorithms for minimizing tracking regret are fixed share and projection update \citep{Herbster1998,Herbster:2001:TBL}, both of which are variants of the MWU method. In each round $t$, both algorithms first compute an intermediate distribution $\w_{t+1}^m$ following MWU in (\ref{eq:mwu-update}). Then, the fixed share algorithm explicitly compels each expert $i \in [K]$ to share a fraction of its assigned distribution with the other experts:
\begin{equation*}
   \w_{t+1}[i] = (1 - \alpha) \w_{t+1}^m [i] + \sum_{j \in [K] : j \neq i} \frac{\alpha}{K-1} \w_{t+1}^m[j].
\end{equation*}
Different from this, in projection update, sharing is implicitly performed by projecting the intermediate distribution $\w_{t+1}^m$ onto a subset of the simplex $\Delta_K$:
\begin{equation}
   \label{def:proj-upda}
   \w_{t+1} = \mathop{\arg\min}_{\w \in \Delta_K \cap [\alpha, 1]^K} \D_{\phi}(\w \Vert \w_{t+1}^m)
\end{equation}
where $\D_{\phi}(\cdot \Vert \cdot)$ denotes Bregman divergence with respect to the negative entropy function and will be made clear in the next section. In both algorithms, the parameter $\alpha$ controls the extent of sharing. \citet{Fixed:Share:NIPS12} showed that with appropriate configuration of parameters $\eta$ and $\alpha$, both algorithms enjoy the first-order tracking regret bound. \citet{pmlr-v40-Luo15} developed the AdaNormalHedge method, which is parameter-free and attains a refined first-order bound on tracking regret. 

\section{Algorithms}
In this section, we first introduce the online mirror descent framework, then propose our algorithms, and finally discuss parameter tuning for our algorithms.
\subsection{Online Mirror Descent}
Our algorithms are developed under the online mirror descent (OMD) framework. OMD is believed to be the gold standard for online learning \citep{srebro2011universality,steinhardt2014adaptivity}, and a variety of algorithms such as online gradient descent and exponentiated gradient can be derived from this framework \citep{Online:suvery}. As outlined in Algorithm \ref{alg:omd}, at each round $t$, after observing the loss vector $\lh_t$, OMD (configured with learning rate $\eta$) updates the distribution as
\begin{equation}
   \label{eq:omd-update}
   \w_{t+1} = \mathop{\arg\min}_{\w \in \Delta_K} \langle \w, \eta \lh_t \rangle + \D_{\phi}(\w \Vert \w_t)
\end{equation}
where $\Delta_K$ is the $K$-simplex:
\begin{equation*}
   \Delta_K = \left\{ \w \in \mathbb{R}^K ~\Big\vert~ \w[i] \geq 0, \forall i \in [K];~ \sum_{i=1}^K \w[i] = 1 \right\}
\end{equation*}
$\phi$ is the negative Shannon entropy function:
\begin{equation}
   \label{def:neg-entr}
   \phi(\w) = \sum_{i=1}^K \w[i] \log{\w[i]}
\end{equation}
and $\D_{\phi}(\cdot \Vert \cdot)$ denotes the Bregman divergence with respect to $\phi$:
\begin{equation}
   \label{def:bregman}
   \D_{\phi}(\x \Vert \y) = \phi(\x) - \phi(\y) - \langle \x-\y, \nabla \phi(\y) \rangle.
\end{equation}

Though seemingly different, Algorithm \ref{alg:omd} is exactly identical to the classic MWU method \citep{Online:suvery}, which can achieve an $O(\sqrt{T\log{K}})$ static regret bound but fails to attain meaningful tracking regret bounds. However, we show that Algorithm \ref{alg:omd} with a simple yet powerful modification---replacing the simplex $\Delta_K$ with a clipped simplex defined below---is able to achieve meaningful tracking regret bounds.
\label{sec:alg}
\begin{algorithm}[t]
   \caption{Online Mirror Descent (specialized for prediction with expert advice)}
   \begin{algorithmic}[1]
   \REQUIRE learning rate $\eta > 0$
   \STATE Initialize $\w_1 = (1/K, 1/K, \ldots, 1/K)$
   \FOR{$t=1,\ldots,T$}
   \STATE Choose distribution $\w_t$
   \STATE Observe loss vector $\lh_t$ and suffer a loss $\langle \w_t, \lh_t \rangle$
   \STATE 
   $\w_{t+1} = \mathop{\arg\min}_{\w \in \Delta_K} \langle \w, \eta \lh_t \rangle +  \D_{\phi}(\w \Vert \w_t)$
   \ENDFOR
   \end{algorithmic}
   \label{alg:omd}
\end{algorithm}
\begin{thm}
   \label{thm:OMD-A}
   Consider the following clipped simplex
   \begin{equation}
      \label{def:clipped-simplex}
      \widetilde \Delta_K = \left\{ \w \in \Delta_K ~\Big \vert~ \w[i] \geq \frac{S}{TK}, \forall i \in [K] \right\} .
   \end{equation}
   Let $\A$ be a variant of Algorithm \ref{alg:omd} that replaces Step $5$ with 
   \begin{equation}
      \label{eq:update-calA}
      \w_{t+1} = \mathop{\arg\min}_{\w \in \widetilde \Delta_K} \langle \w, \eta \lh_t \rangle +  \D_{\phi}(\w \Vert \w_t) .
   \end{equation}
   For $\eta > 0$, the tracking regret of $\A$ satisfies
   \begin{equation*}
      \TR(T, S) \leq \eta T + \frac{S\log{(KT/S)}}{\eta} + S .
   \end{equation*}
   Picking $\eta = \sqrt{\frac{S\log{(KT/S)}}{T}}$ leads to a tracking regret bound of $\O \big( \sqrt{ST\log{(KT/S)}} \big)$.
\end{thm}

In fact, the technique of restricting the feasible set to be the clipped simplex can be shown as a different form of the projection update method \citep{Herbster:2001:TBL} as follows.
\begin{prop}
   \label{prop-one}
   Let $\A$ be the variant of Algorithm \ref{alg:omd} defined in Theorem \ref{thm:OMD-A} and $\B$ be the projection update method defined in (\ref{def:proj-upda}) configured with $\alpha = S/(TK)$. Let $\w_t$ and $\widehat \w_t$ be the distributions chosen in round $t$ by $\A$ and $\B$ respectively. We have $\w_t = \widehat \w_t$ for all $t \in [T]$.
\end{prop}
Nevertheless, directly using clipped simplex in the updating step of OMD avoids the intermediate distribution appearing in projection update method and allows us to follow the analysis framework of OMD. In the following, we combine clipped simplex with existing algorithms that enjoy data-dependent \emph{static} regret bounds to yield new algorithms with data-dependent \emph{tracking} regret bounds.

\subsection{Proposed Algorithms}
Our first algorithm is a variant of the Prod method \citep{Cesa-Bianchi2005}.  While Prod was known to enjoy the second-order static regret bound, we show that equipped with clipped simplex, this method can also attain similar results for tracking regret. Recall that at each round $t$, after observing the loss vector $\lh_t$, Prod performs the following computation to update the distribution:
\begin{equation}
   \label{eq:prod-update}
   \w_{t+1}[i] = \frac{\w_t[i] (1-\eta \lh_t[i])}{\sum_{j=1}^K  \w_t[j] (1-\eta \lh_t[j])}, \, \forall i \in [K] .
\end{equation}
To combine Prod with clipped simplex, we first re-derive the above update in the OMD framework:
\begin{equation*}
   \w_{t+1} = \mathop{\arg\min}_{\w \in \Delta_K} \, \langle \w, -\log{(1 - \eta \lh_t)} \rangle + \D_{\phi} (\w \Vert \w_t)
\end{equation*}
where the $\log(\cdot)$ function is point-wise. Then, we replace the simplex $\Delta_K$ with the clipped simplex $\widetilde \Delta_K$:
\begin{equation}
   \label{eq:pcs-update}
   \w_{t+1} = \mathop{\arg\min} \limits_{\w \in \widetilde \Delta_K} \, \langle \w, -\log{(1 - \eta \lh_t)} \rangle + \D_{\phi} (\w \Vert \w_t) .\mspace{-2mu}
\end{equation}
We name the obtained algorithm as Prod on Clipped Simplex (PCS), which is summarized in Algorithm \ref{alg:PCS} and achieves the second-order bound on tracking regret as follows.

\begin{algorithm}[t]
   \caption{Prod on Clipped Simplex (PCS)}
   \begin{algorithmic}[1]
   \REQUIRE learning rate $\eta \in (0, 1/2]$
   \STATE Initialize $\w_1$ to be arbitrary distribution in $\widetilde \Delta_K$
   \FOR{$t=1,\ldots,T$}
   \STATE Choose distribution $\w_t$
   \STATE Observe loss vector $\lh_t$ and suffer a loss $\langle \w_t, \lh_t \rangle$
   \STATE Update $\w_t$ according to (\ref{eq:pcs-update})
   \ENDFOR
   \end{algorithmic}
   \label{alg:PCS}
\end{algorithm}

\begin{thm}
   \label{thm:PCS}
   For $\eta \in (0, 1/2]$, the tracking regret of PCS satisfies
   \begin{equation*}
      \TR(T, S) \leq \eta L_2 + \frac{S\log{(KT/S)}}{\eta} + \frac{3S}{2}
   \end{equation*}
   where $L_2$ is the sum of squared loss of the best expert and is defined in (\ref{def:L2}). Picking $\eta = \min \left\{ \sqrt{\frac{S\log{(KT/S)}}{L_2}}, \frac{1}{2} \right\}$ leads to a tracking regret bound of $\O \big(\sqrt{SL_2\log{(KT/S)}} + S\log{(KT/S)}\big)$.
\end{thm}
\begin{rmk}
   \normalfont To the best of our knowledge, it is the first time that the second-order bound is derived for tracking the best expert. Compared to the existing first-order bound \citep{Fixed:Share:NIPS12} of $\O\big(\sqrt{SL_1\log{(KT/S)}} + S\log{(KT/S)}\big)$ with $L_1 = \sum_{t=1}^T \lh_t[\EP_t^*]$, our second-order bound can be much smaller for small losses. 
\end{rmk}

\begin{algorithm}[t]
   \caption{Optimistic descent on Clipped Simplex (OCS)}
   \begin{algorithmic}[1]
      \REQUIRE learning rate $\eta > 0$
      \STATE Initialize $\widetilde \w_1$ to be arbitrary distribution in $\widetilde \Delta_K$ and set $\lh_0 = (0, 0, \ldots, 0)$
      \FOR{$t=1,\ldots,T$}
      \STATE Choose distribution $\w_t$ according to (\ref{eq:up1-ocs-alg})
      \STATE Observe loss vector $\lh_t$ and suffer a loss $\langle \w_t, \lh_t \rangle$
      \STATE Update $\widetilde \w_t$ according to (\ref{eq:up2-ocs-alg})
      \ENDFOR
      \end{algorithmic}
   \label{alg:OCS}
\end{algorithm}

Our second algorithm is a variant of the optimistic mirror descent (OptMD) method \citep{pmlr-v23-chiang12,Predictable:COLT:2013}. In OptMD, there exists an auxiliary sequence of distributions  $\widetilde \w_1, \ldots, \widetilde \w_T$, which proceeds in the same way as online mirror descent in (\ref{eq:omd-update}): 
\begin{equation}
   \label{OptMD1}
   \widetilde \w_{t} = \mathop{\arg\min}_{\w \in \Delta_K} \langle \w, \eta \lh_{t-1} \rangle + \D_{\phi}(\w \Vert \widetilde \w_{t-1}) .
\end{equation}
At each round $t$, based on the auxiliary distribution $\widetilde \w_{t}$, OptMD chooses  $\w_t$ as
\begin{equation}
   \label{OptMD2}
   \w_{t} = \mathop{\arg\min}_{\w \in \Delta_K} \langle \w, \eta \lh_{t-1} \rangle + \D_{\phi}(\w \Vert \widetilde \w_{t}) .
\end{equation}
The intuition behind OptMD, as spelled out by \citet{pmlr-v23-chiang12}, is as follows. On one hand, if the loss vectors move slowly (i.e., $\lh_{t}$ is close to $\lh_{t-1}$), the chosen distribution $\w_t$ in (\ref{OptMD2}) can be seen as an approximation to the following imaginary perfect choice:
\begin{equation*}
   \w_t = \mathop{\arg\min}_{\w \in \Delta_K} \langle \w, \eta \lh_{t} \rangle + \D_{\phi}(\w \Vert \widetilde \w_{t})
\end{equation*}
which minimizes the loss of the $t$-th round $\langle \w, \lh_t \rangle$ and thus leads to a small regret. On the other hand, even under the worst case that $\lh_t$ is far away from $\lh_{t-1}$, the Bregman divergence term $\D_{\phi}(\w \Vert \widetilde \w_{t})$ in (\ref{OptMD2}) protects $\w_t$ from deviating too much from $\widetilde \w_t$ in (\ref{OptMD1}) and hence prevents from incurring a large regret. 

While OptMD was originally designed for static regret, we show that by combining with clipped simplex, the above intuition also translates into similar results for tracking regret. Specifically, we replace the simplex $\Delta_K$ in (\ref{OptMD1}) and (\ref{OptMD2}) with the clipped simplex $\widetilde \Delta_K$:
\begin{align}
   \w_{t} & = \mathop{\arg\min} \limits_{\w \in \widetilde \Delta_K} \langle \w, \eta \lh_{t-1} \rangle + \D_{\phi}(\w \Vert \widetilde \w_{t}); \label{eq:up1-ocs-alg} \\
   \widetilde \w_{t+1} & = \mathop{\arg\min}_{\w \in \widetilde \Delta_K} \langle \w, \eta \lh_t \rangle + \D_{\phi}(\w \Vert \widetilde \w_t). \label{eq:up2-ocs-alg}
\end{align}
The resulting algorithm is outlined in Algorithm \ref{alg:OCS}, which is referred to as Optimistic descent on Clipped Simplex (OCS) and enjoys the following path-length bound on tracking regret.
\begin{thm}
   \label{thm:OCS}
   For $\eta > 0$, the tracking regret of OCS satisfies
   \begin{equation*}
      \TR(T, S) \leq \eta P_\infty + \frac{S\log{(KT/S)}}{\eta} + S
   \end{equation*}
   where $P_\infty$ is the sum of squared difference between consecutive loss vectors and is defined in (\ref{def:Pinfty}). Picking $\eta = \sqrt{\frac{S\log{(KT/S)}}{P_\infty}}$ leads to a tracking regret bound of $\O\big(\sqrt{SP_\infty\log{(KT/S)}} + S\big)$.
\end{thm}
\begin{rmk}
   \normalfont
   The above path-length bound is generally not comparable to the second-order bound derived in Theorem \ref{thm:PCS}. Specifically, the path-length bound becomes smaller when the loss vector gradually changes, while the second-order bound is better for small losses.
\end{rmk}
\begin{rmk}
   \normalfont 
   In each round, the main computational overhead of our two algorithms is solving the minimization problems. Thanks to the fact that the clipped simplex $\widetilde \Delta_K$ is a convex set and all the objective functions to minimize are convex, we can solve these minimization problems efficiently by using general convex optimization methods \citep{Convex-Optimization}.
\end{rmk}
\subsection{Parameter Tuning}
We note that to attain the second-order and the path-length tracking regret bounds, our algorithms PCS and OCS require prior knowledge of $L_2$ and $P_\infty$ respectively for tuning the learning rates. Obtaining the second-order bound without such hindsight knowledge is highly challenging due to the non-monotonic issue \citep{gaillard2014second} and remains open even in the context of static regret. 
However, by employing a variant of doubling trick \citep{wei2018more}, we can provide a parameter-free version of OCS achieving the path-length bound.
\begin{algorithm}[t]
   \caption{Optimistic descent on Clipped Simplex plus doubling trick (OCS+)}
   \begin{algorithmic}[1]
      \REQUIRE Time horizon $T$, maximum number of switches $S-1$
      \STATE Initialize $\widetilde \w_1 \in \widetilde \Delta_K$ arbitrarily and set $m = 1, \eta_1 = \sqrt{S\log{(KT/S)}}, \tau_1 = 0, \lh_0 = (0, \ldots, 0)$
      \WHILE{$t \leq T$}
      \STATE $P_m = 0$
      \STATE $t = \tau_m + 1$
      \WHILE{$t \leq T$}
      \STATE Choose distribution $\w_{t} = \mathop{\arg\min}_{\w \in \widetilde \Delta_K} \langle \w, \eta_m \lh_{t-1} \rangle + \D_{\phi}(\w \Vert \widetilde \w_{t})$
      \STATE Observe loss vector $\lh_t$ and suffer a loss $\langle \w_t, \lh_t \rangle$
      \STATE Update
      $  \widetilde \w_{t+1} = \mathop{\arg\min}_{\w \in \widetilde \Delta_K} \langle \w, \eta_m \lh_t \rangle + \D_{\phi}(\w \Vert \widetilde \w_t)  $
      \STATE $P_m = P_m + \Vert \lh_t - \lh_{t-1} \Vert_\infty^2$
      \IF{$\eta_m > \sqrt{\frac{S\log{(KT/S)}}{P_m}}$}
      \STATE $\eta_{m+1} = \eta_m / 2$
      \STATE $\tau_{m+1} = t$
      \STATE $m = m+1$
      \STATE {\bf break}
      \ELSE
      \STATE $t = t+1$
      \ENDIF
      \ENDWHILE
      \ENDWHILE
      \end{algorithmic}
   \label{alg:OCSPlus}
\end{algorithm}

The main idea is to split the time horizon $[1, T]$ into a serials of epochs, and run OCS with different learning rates in different epochs. Specifically, let $m=1,2,\ldots$ index the epoch. We denote the learning rate used in the $m$-th epoch by $\eta_m$ and the starting round of the $m$-th epoch by $\tau_m+1$. For every epoch $m$, we maintain a variable $P_m$, which is initialized to be $0$ in the beginning of the $m$-th epoch and updated in each round $t$ (belonging to this epoch) as
$
   P_m = P_m + \Vert \lh_t - \lh_{t-1} \Vert_\infty^2 .
$
In other words, at the end of round $t$, we have
$
   P_m = \sum_{s=\tau_{m}+1}^t \Vert \lh_s - \lh_{s-1} \Vert_\infty^2
$, which reveals the fact that $P_m$ denotes the path-length (pertaining to the $m$-th epoch) up to round $t$. The role of $P_m$ is as follows: At the end of each round (in the $m$-th epoch), we will check whether the inequality
$
   \eta_m > \sqrt{\frac{S\log{(KT/S)}}{P_m}} 
$
holds true or not. If it is true, we conclude that the currently-used learning rate $\eta_m$  is not suitable and hence enter into a new epoch (the $(m+1)$-th epoch) with half the learning rate:
$
   \eta_{m+1} = \frac{\eta_m}{2}.
$
The above procedure is summarized in Algorithm \ref{alg:OCSPlus}, which is referred to as OCS+ and enjoys the following theoretical guarantee.
\begin{thm}
   \label{thm:OCSD}
   The tracking regret of the OCS+ algorithm satisfies
   \begin{equation*}
      \TR(T, S)  \leq \O\big(\sqrt{S (P_\infty + 1)\log{(KT/S)}} + S \big)
   \end{equation*}
   where $P_\infty$ is defined in (\ref{def:Pinfty}).
\end{thm}

Finally, we would like to discuss the possibility of obtaining parameter-free algorithms with the second-order bound by the two-layer approach \citep{hazan2007adaptive}, where multiple copies of a base algorithm are created in different rounds, and their outputs are combined by a sleeping expert algorithm. While this approach can lead to tracking regret bounds \citep{adamskiy2016closer}, it currently faces difficulties in achieving our second-order bound.

Specifically, let $[p, q] \subseteq [1, T]$ be a time interval and $\E^*$ be the best expert in this interval. Then the regret of the sleeping expert algorithm in $[p, q]$ with respect to $\E^*$ can be decomposed as
\begin{equation*}
   \begin{split}
      & \sum_{t=p}^q \langle \w_t, \lh_t \rangle - \sum_{t=p}^q \lh_t[\E^*] \\
      = & \underbrace{\sum_{t=p}^q \langle \w_t, \lh_t \rangle -  \sum_{t=p}^q \langle \w_t^p, \lh_t \rangle}_{\mathcal{A}}
      +  \underbrace{\sum_{t=p}^q \langle \w_t^p, \lh_t \rangle - \sum_{t=p}^q \lh_t[\E^*]}_{\mathcal{B}}
   \end{split}
\end{equation*}
where $\w_t$ and $\w_t^p$ denote the outputs in round $t$ of the sleeping expert algorithm and the base algorithm created in round $p$, respectively. The term $\mathcal{A}$ is the regret of the sleeping expert algorithm  with respect to the base algorithm, and the term $\mathcal{B}$ is the regret of the base algorithm with respect to $\E^*$. 

To obtain the second-order tracking regret bound, one needs to derive second-order bounds for both $\A$ and $\B$. While for $\B$ this can be easily done by picking Prod as the base algorithm, unfortunately, for $\A$ no existing sleeping expert algorithms can achieve our second-order bound. In fact, the state-of-the-art bounds for the sleeping expert problem are a refined first-order bound \citep{pmlr-v40-Luo15} and a second-order \emph{excess loss} bound \citep{gaillard2014second}, the latter of which depends on the sum of squared \emph{excess loss} (i.e., the difference between the loss of the learner and that of the best expert) and is hence not comparable to our second-order bound that depends on the sum of squared loss of the best expert.

\section{Extension to Online Matrix Prediction}
We now extend our algorithms to online matrix prediction \citep{hazan2012near}, which can model a variety of problems such as online collaborative filtering and online max-cut. Before describing the setup, we first introduce some useful definitions and notations. Let $A$ be a $K \times K$ matrix,\footnote{Throughout this section, all matrices are assumed to be symmetric and real.} we use $\Vert A \Vert$ and $\Vert A \Vert_*$ to denote the nuclear and the spectral norms of $A$ respectively, which are defined by
\begin{equation*}
   \Vert A \Vert = \sum_{i=1}^K \vert \lambda_i(A) \vert ;~~~~~ \Vert A \Vert_* = \max_{i \in [K]} \vert \lambda_i(A) \vert
\end{equation*}
where $\lambda_i(A)$ is the $i$-th eigenvalue of $A$. It is well known that the nuclear norm is the dual norm of the spectral norm, and vice versa. We use $I_K$ to denote the $K \times K$ identity matrix. In matrix settings, the counterpart of the $K$-simplex $\Delta_K$ is the $K$-spectraplex, defined as
\begin{equation*}
   \Omega_K = \Big\{ W \in \mathbb{S}_{+}^K \vert \tr(W) = 1 \Big\}
\end{equation*}
where $\mathbb{S}_{+}^K$ is the set comprised of all $K \times K$ positive semidefinite matrices, and $\tr(\cdot)$ denotes the trace.
Given a matrix $W \in \Omega_K$, let $W=V \Lambda V^T$ be the eigendecomposition of $W$, where $V$ is an orthogonal matrix whose columns are the eigenvectors of $W$, and $\Lambda$ is a diagonal matrix whose
   entries are the eigenvalues of $W$. We define $\log{\Lambda}$ to be a diagonal matrix with $(\log{\Lambda})_{ii} = \log{(\Lambda_{ii})}$ 
   and define $\log{W}$ by
\begin{equation*}
   \log{W} = V(\log{\Lambda})V^T.
\end{equation*}

We are now ready to describe the setup of online matrix prediction, which is taken from \citet{steinhardt2014adaptivity}: In each round $t$, a learner chooses a prediction matrix $W_t \in \Omega_K$, and meanwhile an adversary decides a loss matrix $Z_t$ satisfying $\Vert Z_t \Vert_{*} \leq 1$. Then, the learner observes the loss matrix $Z_t$ and suffers a loss $\tr(W_tZ_t)$. Similarly to (\ref{eq:def:tr}), we define the tracking regret as
\begin{equation*}
   \begin{split}
      \TR(T, S)  = \sum_{t=1}^T \tr(W_tZ_t) - \mspace{-15mu} \min_{(U_1, \ldots, U_T) \in \U(T, S)} \sum_{t=1}^T \tr(U_tZ_t)
       = \sum_{t=1}^T \tr(W_tZ_t) - \sum_{t=1}^T \tr(U_t^* Z_t)
   \end{split}
\end{equation*}
where $\U(T, S)$ is the set of sequences of matrices in $\Omega_K$ with switches not more than $S - 1$:
\begin{equation*}
   \begin{split}
      \U(T, S) = \left\{ (U_1, \ldots, U_T) \in \Omega_K^T  ~\vert~  \sum_{t=2}^T \mathds{1}{\{U_t \neq U_{t-1}\}} \leq S - 1 \right\} 
   \end{split}
\end{equation*}
and $U_1^*,\ldots,U_T^*$ is the best sequence in $\U(T, S)$:
\begin{equation*}
   (U_1^*,\ldots,U_T^*) = \mathop{\arg\min}_{(U_1, \ldots, U_T) \in \U(T, S)} \sum_{t=1}^T \tr(U_tZ_t) .
\end{equation*}
\begin{algorithm}[t]
   \caption{Prod on Clipped SPectraplex (PCSP)}
   \begin{algorithmic}[1]
   \REQUIRE learning rate $\eta \in (0, 1/2]$
   \STATE Initialize $W_1$ to be arbitrary matrix in $\widetilde \Omega_K$
   \FOR{$t=1,\ldots,T$}
   \STATE Choose prediction matrix $W_t$
   \STATE Observe loss matrix $Z_t$ and suffer a loss $\tr(W_tZ_t)$
   \STATE 
   $  W_{t+1} = \mathop{\arg\min}_{W \in \widetilde \Omega_K} \, \tr(-W\log{(I_K - \eta Z_t)}) + \D_{\psi} (W \Vert W_t)$
   \ENDFOR
   \end{algorithmic}
   \label{alg:PCSP}
\end{algorithm}

As pointed out by \citet{steinhardt2014adaptivity}, the problem of prediction with expert advice can be viewed as a special case of online matrix prediction by setting $W_t = \diag(\w_t)$ and $Z_t = \diag(\lh_t)$, where $\diag(\cdot)$ denotes the diagonalization of a vector. Based on this observation, we construct the clipped spectraplex $\widetilde \Omega_K$ as a natural extension of the clipped simplex:
\begin{equation}
   \label{def:clipped-spx}
   \widetilde \Omega_K = \left\{ W \in \mathbb{S}_{+}^K \vert \tr(W) = 1, \lambda_{\min}(W) \geq \frac{S}{TK} \right\}
\end{equation}
where $\lambda_{\min}(W)$ denotes the minimum eigenvalue of $W$. By the Weyl's inequality \citep{weyl1912asymptotische}, it is easy to show that $\widetilde \Omega_K$ is a convex set.
Furthermore, we realize that in matrix algebra, $\tr(AB)$ plays a similar role as $\langle \a, \b \rangle$ for vectors $\a$ and $\b$ \citep{tsuda2005matrix}, and introduce the negative Von Neumann entropy generalizing the negative Shannon entropy:
\begin{equation}
   \psi(W) = \tr(W\log{W}), ~W \in \Omega_K .
\end{equation}
Finally, the Bregman divergence can also be smoothly extended to the matrix function $\psi$:
\begin{equation}
   \label{def:breg-mtx}
   \D_{\psi} (A \Vert B) = \psi(A) - \psi(B) - \tr\big( (A-B)\nabla \psi(B) \big).
\end{equation}

Equipped with these, extending our algorithms to online matrix prediction is straightforward. For brevity, we only provide the extension of our first algorithm PCS in Algorithm \ref{alg:PCSP} (referred to as Prod on Clipped SPectraplex, PCSP), and the extension of our second algorithm can be done in the same way. Similarly to Theorem \ref{thm:PCS}, we have the following theoretical guarantee for PCSP.
\begin{thm}
   \label{thm:PCSP}
   For $\eta \in (0, 1/2]$, the tracking regret of PCSP satisfies
   \begin{equation*}
      \TR(T, S)  \leq \eta M_2 + \frac{S\log{(KT/S)}}{\eta} + \frac{5S}{2}
   \end{equation*}
   where we define
$
      M_2 = \sum_{t=1}^T \tr\big(U_t^* Z_t^2 \big) .
$
   Picking $\eta = \min \left\{ \sqrt{\frac{S\log{(KT/S)}}{M_2}}, \frac{1}{2} \right\}$ leads to a tracking regret bound of $\O\big( \sqrt{SM_2 \log{(KT/S)}} + S\log{(KT/S)} \big)$.
\end{thm}
\begin{rmk}
   \normalfont
   While there exist data-independent tracking regret bounds for the problem of online matrix prediction \citep{pmlr-v48-gyorgy16}, to the best of our knowledge, the bound in Theorem \ref{thm:PCSP} is the first data-dependent bound on tracking regret for this problem.
\end{rmk}

\section{Conclusion and Future Work}
In this paper, we develop two adaptive and efficient algorithms that enjoy data-dependent bounds for the problem of tracking the best expert. The first algorithm is inspired by the Prod algorithm and attains the second-order tracking regret bound improving previous first-order bounds. The second algorithm draws inspiration from the optimistic mirror descent method and achieves the path-length bound offering advantages in slowly moving environments. We also provide an extension of our algorithms and analysis to the problem of online matrix prediction and present the first data-dependent tracking regret bound for this problem.

There are several future directions to pursue. First, in the current study, both the time horizon $T$ and the maximum number of switches $S-1$ are assumed to be known in advance. In the future, we will try to develop more adaptive algorithms that are efficient and can adapt to unknown $T$ and $S$. Second, in the context of static regret, \citet{steinhardt2014adaptivity} have derived a bound on the order of $O(\sqrt{P_*\log{K}} + \log{K})$, where $P_* = \sum_{t=1}^T (\lh_t[\EP_*] - \lh_{t-1}[\EP_*])^2$ and $\EP_*$ is the best expert over $T$ rounds. This bound is better than both the second-order and the path-length bounds. It is appealing to obtain similar results for tracking regret. Finally, in light of recent advances in obtaining data-dependent static regret bounds for the multi-armed bandits problem \citep{wei2018more,bubeck2019improved}, it would be interesting to examine whether  our algorithms and analysis can be extended to the bandits setting.

\bibliography{ref}

\begin{thebibliography}{34}
\providecommand{\natexlab}[1]{#1}
\providecommand{\url}[1]{\texttt{#1}}
\expandafter\ifx\csname urlstyle\endcsname\relax
  \providecommand{\doi}[1]{doi: #1}\else
  \providecommand{\doi}{doi: \begingroup \urlstyle{rm}\Url}\fi

\bibitem[Adamskiy et~al.(2016)Adamskiy, Koolen, Chernov, and
  Vovk]{adamskiy2016closer}
Dmitry Adamskiy, Wouter~M Koolen, Alexey Chernov, and Vladimir Vovk.
\newblock A closer look at adaptive regret.
\newblock \emph{Journal of Machine Learning Research}, 17\penalty0
  (1):\penalty0 706--726, 2016.

\bibitem[Bousquet and Warmuth(2002)]{bousquet2002tracking}
Olivier Bousquet and Manfred~K Warmuth.
\newblock Tracking a small set of experts by mixing past posteriors.
\newblock \emph{Journal of Machine Learning Research}, 3\penalty0
  (Nov):\penalty0 363--396, 2002.

\bibitem[Boyd and Vandenberghe(2004)]{Convex-Optimization}
Stephen Boyd and Lieven Vandenberghe.
\newblock \emph{Convex Optimization}.
\newblock Cambridge University Press, 2004.

\bibitem[Bubeck et~al.(2019)Bubeck, Li, Luo, and Wei]{bubeck2019improved}
S{\'e}bastien Bubeck, Yuanzhi Li, Haipeng Luo, and Chen-Yu Wei.
\newblock Improved path-length regret bounds for bandits.
\newblock In \emph{Proceedings of the 32nd Conference on Learning Theory},
  pages 508--528, 2019.

\bibitem[Cesa-Bianchi and Lugosi(2006)]{bianchi-2006-prediction}
Nicol\`{o} Cesa-Bianchi and G{\'a}bor Lugosi.
\newblock \emph{Prediction, Learning, and Games}.
\newblock Cambridge University Press, 2006.

\bibitem[Cesa-Bianchi et~al.(2005)Cesa-Bianchi, Mansour, and
  Stoltz]{Cesa-Bianchi2005}
Nicol{\`o} Cesa-Bianchi, Yishay Mansour, and Gilles Stoltz.
\newblock Improved second-order bounds for prediction with expert advice.
\newblock In \emph{Proceedings of the 18th Annual Conference on Learning
  Theory}, pages 217--232, 2005.

\bibitem[Cesa-bianchi et~al.(2012)Cesa-bianchi, Gaillard, Lugosi, and
  Stoltz]{Fixed:Share:NIPS12}
Nicol\`{o} Cesa-bianchi, Pierre Gaillard, Gabor Lugosi, and Gilles Stoltz.
\newblock Mirror descent meets fixed share (and feels no regret).
\newblock In \emph{Advances in Neural Information Processing Systems 25}, pages
  980--988, 2012.

\bibitem[Chiang et~al.(2012)Chiang, Yang, Lee, Mahdavi, Lu, Jin, and
  Zhu]{pmlr-v23-chiang12}
Chao-Kai Chiang, Tianbao Yang, Chia-Jung Lee, Mehrdad Mahdavi, Chi-Jen Lu, Rong
  Jin, and Shenghuo Zhu.
\newblock Online optimization with gradual variations.
\newblock In \emph{Proceedings of the 25th Annual Conference on Learning
  Theory}, pages 6.1--6.20, 2012.

\bibitem[Freund and Schapire(1997)]{FREUND1997119}
Yoav Freund and Robert~E Schapire.
\newblock A decision-theoretic generalization of on-line learning and an
  application to boosting.
\newblock \emph{Journal of Computer and System Sciences}, 55\penalty0
  (1):\penalty0 119 -- 139, 1997.

\bibitem[Gaillard et~al.(2014)Gaillard, Stoltz, and
  Van~Erven]{gaillard2014second}
Pierre Gaillard, Gilles Stoltz, and Tim Van~Erven.
\newblock A second-order bound with excess losses.
\newblock In \emph{Proceedings of The 27th Conference on Learning Theory},
  pages 176--196, 2014.

\bibitem[Golden(1965)]{golden1965lower}
Sidney Golden.
\newblock Lower bounds for the helmholtz function.
\newblock \emph{Physical Review}, 137\penalty0 (4B):\penalty0 B1127, 1965.

\bibitem[Gyorgy and Szepesvari(2016)]{pmlr-v48-gyorgy16}
Andras Gyorgy and Csaba Szepesvari.
\newblock Shifting regret, mirror descent, and matrices.
\newblock In \emph{Proceedings of The 33rd International Conference on Machine
  Learning}, pages 2943--2951, 2016.

\bibitem[Hannan(1957)]{hannan1957approximation}
James Hannan.
\newblock Approximation to bayes risk in repeated play.
\newblock \emph{Contributions to the Theory of Games}, 3:\penalty0 97--139,
  1957.

\bibitem[Hazan and Kale(2010)]{variation:bound}
Elad Hazan and Satyen Kale.
\newblock Extracting certainty from uncertainty: regret bounded by variation in
  costs.
\newblock \emph{Machine Learning}, 80\penalty0 (2-3):\penalty0 165--188, 2010.

\bibitem[Hazan and Seshadhri(2007)]{hazan2007adaptive}
Elad Hazan and Comandur Seshadhri.
\newblock Adaptive algorithms for online decision problems.
\newblock In \emph{Electronic Colloquium on Computational Complexity},
  volume~14, 2007.

\bibitem[Hazan et~al.(2012)Hazan, Kale, and Shalev-Shwartz]{hazan2012near}
Elad Hazan, Satyen Kale, and Shai Shalev-Shwartz.
\newblock Near-optimal algorithms for online matrix prediction.
\newblock In \emph{Proceedings of the 25th Annual Conference on Learning
  Theory}, pages 38--1, 2012.

\bibitem[Herbster and Warmuth(1998)]{Herbster1998}
Mark Herbster and Manfred~K. Warmuth.
\newblock Tracking the best expert.
\newblock \emph{Machine Learning}, 32\penalty0 (2):\penalty0 151--178, 1998.

\bibitem[Herbster and Warmuth(2001)]{Herbster:2001:TBL}
Mark Herbster and Manfred~K. Warmuth.
\newblock Tracking the best linear predictor.
\newblock \emph{Journal of Machine Learning Research}, 1:\penalty0 281--309,
  2001.

\bibitem[Kalai and Vempala(2003)]{kalai2003efficient}
Adam Kalai and Santosh Vempala.
\newblock Efficient algorithms for online decision problems.
\newblock In \emph{Proceedings of the 16th Annual Conference on Learning
  Theory}, pages 26--40. 2003.

\bibitem[Kivinen and Warmuth(1997)]{kivinen1997exponentiated}
Jyrki Kivinen and Manfred~K Warmuth.
\newblock Exponentiated gradient versus gradient descent for linear predictors.
\newblock \emph{Information and Computation}, 132\penalty0 (1):\penalty0 1--63,
  1997.

\bibitem[Littlestone and Warmuth(1994)]{LITTLESTONE1994212}
Nick Littlestone and Manfred~K. Warmuth.
\newblock The weighted majority algorithm.
\newblock \emph{Information and Computation}, 108\penalty0 (2):\penalty0
  212--261, 1994.

\bibitem[Luo and Schapire(2015)]{pmlr-v40-Luo15}
Haipeng Luo and Robert~E. Schapire.
\newblock Achieving all with no parameters: Adanormalhedge.
\newblock In \emph{Proceedings of The 28th Conference on Learning Theory},
  pages 1286--1304, 2015.

\bibitem[Rakhlin and Sridharan(2013)]{Predictable:COLT:2013}
Alexander Rakhlin and Karthik Sridharan.
\newblock Online learning with predictable sequences.
\newblock In \emph{Proceedings of the 26th Conference on Learning Theory},
  pages 993--1019, 2013.

\bibitem[Shalev-Shwartz(2007)]{Shai:thesis}
Shai Shalev-Shwartz.
\newblock \emph{Online Learning: Theory, Algorithms, and Applications}.
\newblock PhD thesis, The Hebrew University of Jerusalem, 2007.

\bibitem[Shalev-Shwartz(2011)]{Online:suvery}
Shai Shalev-Shwartz.
\newblock Online learning and online convex optimization.
\newblock \emph{Foundations and Trends in Machine Learning}, 4\penalty0
  (2):\penalty0 107--194, 2011.

\bibitem[Srebro et~al.(2011)Srebro, Sridharan, and
  Tewari]{srebro2011universality}
Nati Srebro, Karthik Sridharan, and Ambuj Tewari.
\newblock On the universality of online mirror descent.
\newblock In \emph{Advances in neural information processing systems 24}, pages
  2645--2653, 2011.

\bibitem[Steinhardt and Liang(2014)]{steinhardt2014adaptivity}
Jacob Steinhardt and Percy Liang.
\newblock Adaptivity and optimism: An improved exponentiated gradient
  algorithm.
\newblock In \emph{Proceedings of the 31st International Conference on Machine
  Learning}, pages 1593--1601, 2014.

\bibitem[Thompson(1965)]{thompson1965inequality}
Colin~J Thompson.
\newblock Inequality with applications in statistical mechanics.
\newblock \emph{Journal of Mathematical Physics}, 6\penalty0 (11):\penalty0
  1812--1813, 1965.

\bibitem[Tsuda et~al.(2005)Tsuda, R{\"a}tsch, and Warmuth]{tsuda2005matrix}
Koji Tsuda, Gunnar R{\"a}tsch, and Manfred~K Warmuth.
\newblock Matrix exponentiated gradient updates for on-line learning and
  bregman projection.
\newblock \emph{Journal of Machine Learning Research}, 6\penalty0
  (Jun):\penalty0 995--1018, 2005.

\bibitem[Vovk(1999)]{Vovk1999}
V.~Vovk.
\newblock Derandomizing stochastic prediction strategies.
\newblock \emph{Machine Learning}, 35\penalty0 (3):\penalty0 247--282, 1999.

\bibitem[Vovk(1990)]{Vovk1990Aggregating}
Volodimir~G. Vovk.
\newblock Aggregating strategies.
\newblock In \emph{Proceedings of the 3rd Annual Workshop on Computational
  Learning Theory}, pages 371--386, 1990.

\bibitem[Wei and Luo(2018)]{wei2018more}
Chen-Yu Wei and Haipeng Luo.
\newblock More adaptive algorithms for adversarial bandits.
\newblock In \emph{Proceedings of the 31st Conference On Learning Theory},
  pages 1263--1291, 2018.

\bibitem[Wei et~al.(2016)Wei, Hong, and Lu]{wei2016tracking}
Chen-Yu Wei, Yi-Te Hong, and Chi-Jen Lu.
\newblock Tracking the best expert in non-stationary stochastic environments.
\newblock In \emph{Advances in Neural Information Processing Systems 29}, pages
  3972--3980, 2016.

\bibitem[Weyl(1912)]{weyl1912asymptotische}
Hermann Weyl.
\newblock Das asymptotische verteilungsgesetz der eigenwerte linearer
  partieller differentialgleichungen (mit einer anwendung auf die theorie der
  hohlraumstrahlung).
\newblock \emph{Mathematische Annalen}, 71\penalty0 (4):\penalty0 441--479,
  1912.

\end{thebibliography}
\newpage
\appendix

\section{Proof of Proposition \ref{prop-one}}
We prove the statement $\w_t = \widehat \w_t, ~\forall t \in [T]$ by mathematical induction.

(i) $\w_1 = \widehat \w_1$ holds trivially as both are equal to $(1/K, 1/K, \ldots, 1/K)$.

(ii) Suppose $\w_t = \widehat \w_t$ holds for some $t \geq 1$. We show that the statement is also true for $t+1$. First, we state the expression of $\widehat \w_{t+1}$ according to (\ref{def:proj-upda})
and (\ref{eq:mwu-update}):
\begin{align}
   \widehat \w_{t+1}^m[i]  & = \frac{\widehat \w_t[i] \exp(-\eta \lh_t[i])}{\sum_{j=1}^K  \widehat \w_t[j] \exp(-\eta \lh_t[j])}, \, \forall i \in [K]; \label{eq:pf-prop-up0} \\
   \widehat \w_{t+1} & = \mathop{\arg\min}_{\w \in \Delta_K \cap [\alpha, 1]^K} \D_{\phi}(\w \Vert \widehat \w_{t+1}^m) \label{eq:pf-prop-up1} .
\end{align}
Note that for $\alpha = S/(TK)$, we have $\Delta_K \cap [\alpha, 1]^K = \widetilde \Delta_K$. 
Thus, (\ref{eq:pf-prop-up1}) can rewritten as
\begin{equation}
   \label{eq:pf-prop-up3}
   \widehat \w_{t+1} = \mathop{\arg\min}_{\w \in \widetilde \Delta_K } \D_{\phi}(\w \Vert \widehat \w_{t+1}^m) .
\end{equation}
For clarity, we here also restate $\w_{t+1}$, which is defined in (\ref{eq:update-calA}):
\begin{equation}
   \label{eq:pf-prop-cA}
   \w_{t+1} = \mathop{\arg\min}_{\w \in \widetilde \Delta_K} \langle \w, \eta \lh_t \rangle +  \D_{\phi}(\w \Vert \w_t) .
\end{equation}
To proceed, we define a convex function on $\widetilde \Delta_K$:
\begin{equation}
   \label{eq:porp-def-f}
   f(\w) = \langle \w, \eta \lh_t \rangle +  \D_{\phi}(\w \Vert \w_t), ~ \w \in \widetilde \Delta_K .
\end{equation}
By (\ref{eq:pf-prop-cA}) and (\ref{eq:pf-prop-up3}), we have $\w_{t+1} = \mathop{\arg\min}_{\w \in \widetilde \Delta_K} f(\w)$ and $\widehat \w_{t+1} \in \widetilde \Delta_K$, which implies
\begin{equation}
   \label{eq:prop-pf-gt}
   f(\w_{t+1}) \leq f(\widehat \w_{t+1}) .
\end{equation}
It remains to show that the opposite, i.e., $f(\widehat \w_{t+1}) \leq f( \w_{t+1})$, also holds. To this end, we introduce the following lemma.
\begin{lma}
   \label{lm-prop-1}
   Let $\u \in \mathbb{R}^K$ be any $K$-dimensional vector satisfying $\sum_{i=1}^K \u[i] = 1$. We have
   \begin{equation*}
      \langle \u - \widehat \w_{t+1}^m , \eta \lh_t + \nabla \phi(\widehat \w_{t+1}^m) - \nabla \phi (\widehat \w_t) \rangle = 0 .
   \end{equation*}
\end{lma}
Consider $\u =  \widehat \w_{t+1}^m +  \widehat \w_{t+1} - \w_{t+1}$. We have
\begin{equation*}
   \sum_{i=1}^K \u[i] = \sum_{i=1}^K (\widehat \w_{t+1}^m + \widehat \w_{t+1}  -  \w_{t+1})[i] =  \sum_{i=1}^K \widehat \w_{t+1}^m[i] + \sum_{i=1}^K \widehat \w_{t+1}[i]  -  \sum_{i=1}^K  \w_{t+1}[i] = 1 .
\end{equation*}
Thus, we can apply Lemma \ref{lm-prop-1} and get
\begin{equation}
   \label{eq:pf-prop-equ}
   \langle \widehat \w_{t+1} -  \w_{t+1}, \eta \lh_t + \nabla \phi(\widehat \w_{t+1}^m) - \nabla \phi (\widehat \w_t) \rangle = 0 .
\end{equation}
On the other hand, note that $\widetilde \Delta_K$ is a convex set. By (\ref{eq:pf-prop-up3}) and the first order optimal condition, we have
\begin{equation*}
   \langle \widehat \w_{t+1} - \v, \nabla \phi(\widehat \w_{t+1}) - \nabla \phi (\widehat \w_{t+1}^m) \rangle \leq 0, ~\forall \v \in \widetilde \Delta_K .
\end{equation*}
By (\ref{eq:pf-prop-cA}), $\w_{t+1} \in \widetilde \Delta_K$. Therefore, we can substitute $\v = \w_{t+1}$ into the above inequality and obtain
\begin{equation}
   \label{eq:pf-prop-inequ}
   \langle \widehat \w_{t+1} - \w_{t+1}, \nabla \phi(\widehat \w_{t+1}) - \nabla \phi (\widehat \w_{t+1}^m) \rangle \leq 0 .
\end{equation}
Adding (\ref{eq:pf-prop-equ}) to (\ref{eq:pf-prop-inequ}) gives
\begin{equation*}
   \langle \widehat \w_{t+1} - \w_{t+1}, \eta \lh_t + \nabla \phi(\widehat \w_{t+1}) - \nabla \phi (\widehat \w_t) \rangle \leq 0 .
\end{equation*}
By the definition of $f$ in (\ref{eq:porp-def-f}) and the assumption $\w_t = \widehat \w_t$, we have $ \eta \lh_t + \nabla \phi(\widehat \w_{t+1}) - \nabla \phi (\widehat \w_t) =\nabla f(\widehat \w_{t+1})$. Combining this with the above inequality and noticing that $f$ is convex, we get
\begin{equation}
   \label{eq:prop-pf-lt}
   f(\widehat \w_{t+1}) - f(\w_{t+1}) \leq \langle \widehat \w_{t+1} - \w_{t+1}, \nabla f(\widehat \w_{t+1}) \rangle \leq 0.
\end{equation}
Combining (\ref{eq:prop-pf-gt}) and (\ref{eq:prop-pf-lt}) and recalling $\w_{t+1} = \mathop{\arg\min}_{\w \in \widetilde \Delta_K} f(\w)$, we obtain $f(\widehat \w_{t+1}) = f(\w_{t+1}) = \min_{\w \in \widetilde \Delta_K} f(\w) $. Finally, since $\phi$ and hence $f$ are strongly convex functions, $f(\widehat \w_{t+1}) = f(\w_{t+1}) = \min_{\w \in \widetilde \Delta_K} f(\w)$ implies $\widehat \w_{t+1} = \w_{t+1}$.

\section{Proof of Theorem \ref{thm:OMD-A}}
\label{sec:pf-omd-th1}
By the definition of tracking regret in (\ref{eq:def:tr})--(\ref{def:best-sequence-of-experts}), we can divide the time horizon $[1, T]$ into $S$ disjoint intervals $[\I_1, \I_2), \ldots, [\I_S, \I_{S+1})$ with $\I_1 = 1$ and $\I_{S+1} = T+1$ such that in each interval $[\I_s, \I_{s+1}), s \in [S]$, the compared expert $\EP_t^*$ remains the same, i.e., 
\begin{equation}
   \label{eq:samee}
   \mathcal{E}_{\I_s}^* = \EP_{\I_s+1}^* = \EP_{\I_s+2}^* =  \cdots = \EP_{\I_{s+1} - 1}^*, ~\forall s \in [S] .
\end{equation}
Fix $s \in [S]$. We now consider the tracking regret in the $s$-th interval $[\I_s, \I_{s+1})$:
\begin{equation*}
   \sum_{t = \I_s}^{\I_{s+1}-1} \big( \hat \ell_t - \lh_t [\EP_t^*] \big) = \sum_{t = \I_s}^{\I_{s+1}-1} \big( \langle \w_t, \lh_t \rangle - \lh_t [\EP_t^*] \big).
\end{equation*}
To express the term $\lh_t [\EP_t^*]$ as an inner product between two vectors, we introduce one-hot vectors $\e_1, \ldots, \e_T$ defined as
\begin{equation}
   \e_t[i] = \begin{cases}
      1, & i = \EP_t^* \cr
      0, & \text{otherwise}
   \end{cases}, ~~ \forall i \in [K].
\end{equation}
Then, we have $\lh_t [\EP_t^*] = \langle \e_t, \lh_t \rangle$ and
\begin{equation}
   \label{eq:tr-s-2}
   \sum_{t = \I_s}^{\I_{s+1}-1} \big( \langle \w_t, \lh_t \rangle - \lh_t [\EP_t^*] \big) = \sum_{t = \I_s}^{\I_{s+1}-1} \langle \w_t - \e_t, \lh_t \rangle.
\end{equation}
We further define $\bar \e_t \in \widetilde \Delta_K$ by
\begin{equation}
   \label{def:bar-e}
   \bar \e_t[i] = (1-\frac{S}{T}) \e_t[i] + \frac{S}{TK}, ~\forall i \in [K]
\end{equation}
and decompose the right-hand side of (\ref{eq:tr-s-2}) as
\begin{equation*}
   \sum_{t = \I_s}^{\I_{s+1}-1} \langle \w_t - \e_t, \lh_t \rangle = \sum_{t = \I_s}^{\I_{s+1}-1} \langle \w_t - \bar \e_t, \lh_t \rangle + \sum_{t = \I_s}^{\I_{s+1}-1} \langle \bar \e_t - \e_t, \lh_t \rangle
\end{equation*}
where the last term can be bounded by the following lemma.
\begin{lma}
   \label{lm:bare-e-lt}
   For any $\lh_t \in [0, 1]^K$ and any $\e_t \in \Delta_K$, let $\bar \e_t$ be defined as in (\ref{def:bar-e}). We have
   \begin{equation*}
      \langle \bar \e_t - \e_t, \lh_t \rangle \leq \frac{S}{T} .
   \end{equation*}
\end{lma}
It follows that
\begin{align}
      \TR(T,S)  \nonumber 
      =  \sum_{s=1}^S \sum_{t = \I_s}^{\I_{s+1}-1} \big( \hat \ell_t - \lh_t [\EP_t^*] \big) & = \sum_{s=1}^S \sum_{t = \I_s}^{\I_{s+1}-1} \langle \w_t - \e_t, \lh_t \rangle  \nonumber \\ 
      & \leq \sum_{s=1}^S \sum_{t = \I_s}^{\I_{s+1}-1} \langle \w_t - \bar \e_t, \lh_t \rangle + \sum_{s=1}^S \sum_{t = \I_s}^{\I_{s+1}-1} \frac{S}{T} \label{pf-tr-omd-sp-1} \\ 
      & = \sum_{s=1}^S \sum_{t = \I_s}^{\I_{s+1}-1} \langle \w_t - \bar \e_t, \lh_t \rangle + S \nonumber .
\end{align}
Then, we decompose $\langle \w_t - \bar \e_t, \lh_t \rangle$ as
\begin{equation}
   \label{eq:pf-thm1:dec}
   \langle \w_t - \bar \e_t, \lh_t \rangle = \langle \w_t - \w_{t+1}, \lh_t \rangle + \langle \w_{t+1} - \bar \e_t, \lh_t \rangle .
\end{equation}
The term $\langle \w_t - \w_{t+1}, \lh_t \rangle$ can be bounded by the following lemma.
\begin{lma}
   \label{lm:pf-thm1:w-t-t1}
   For any $t \in [T]$, we have
   \begin{equation}
      \label{eq:pf-thm1:w-t-t1}
      \langle \w_t - \w_{t+1}, \lh_t \rangle \leq \eta.
   \end{equation}
\end{lma}
It remains to bound the term $\langle \w_{t+1} - \bar \e_t, \lh_t \rangle$. To this end, we define a convex function on the clipped simplex $\widetilde \Delta_K$:
\begin{equation*}
   f(\w) =  \langle \w, \eta \lh_t \rangle +  \D_{\phi}(\w \Vert \w_t), ~\w \in \widetilde \Delta_K
\end{equation*}
and rewrite the updating step in (\ref{eq:update-calA}) as
\begin{equation*}
   \w_{t+1} = \mathop{\arg\min}_{\w \in \widetilde \Delta_K} f(\w) .
\end{equation*}
By the first order optimal condition, we have
\begin{equation*}
   \langle \w_{t+1} - \u, \nabla f(\w_{t+1}) \rangle \leq 0, ~\forall \u \in \widetilde \Delta_K .
\end{equation*}
Substituting $\u = \bar \e_t$, we get
\begin{equation*}
   \begin{split}
      \langle \w_{t+1} - \bar \e_t, \nabla f(\w_{t+1}) \rangle & \leq 0 \\
      \langle \w_{t+1} - \bar \e_t, \eta \lh_t + \nabla \phi(\w_{t+1}) - \nabla \phi (\w_t)) \rangle & \leq 0 \\
      \eta \langle \w_{t+1} - \bar \e_t, \lh_t \rangle & \leq \langle \bar \e_t - \w_{t+1}, \nabla \phi(\w_{t+1}) - \nabla \phi (\w_t) \rangle .
   \end{split}
\end{equation*}
Thus, we have
\begin{equation*}
   \begin{split}
      \langle \w_{t+1} - \bar \e_t, \lh_t \rangle & \leq \frac{1}{\eta} \langle \bar \e_t, \nabla \phi(\w_{t+1}) - \nabla \phi (\w_t) \rangle - \frac{1}{\eta} \langle \w_{t+1}, \nabla \phi(\w_{t+1}) - \nabla \phi (\w_t) \rangle \\
      & = \frac{1}{\eta} \langle \bar \e_t, \nabla \phi(\w_{t+1}) - \nabla \phi (\w_t) \rangle - \frac{1}{\eta} \D_{\phi} (\w_{t+1} \Vert \w_t) \\
      & \leq \frac{1}{\eta} \langle \bar \e_t, \nabla \phi(\w_{t+1}) - \nabla \phi (\w_t) \rangle
   \end{split}
\end{equation*}
where the first equality follows from the definition of Bregman divergence in (\ref{def:bregman}), and the last inequality holds since Bregman divergence is always non-negative. Summing the above inequality over $t=\I_s,\ldots,\I_{s+1}-1$, we get
\begin{equation}
   \label{eq:pf-thm1-wt1e}
   \begin{split}
      \sum_{t=\I_s}^{\I_{s+1}-1} \langle \w_{t+1} - \bar \e_t, \lh_t \rangle & \leq \frac{1}{\eta}  \sum_{t=\I_s}^{\I_{s+1}-1} \langle \bar \e_t, \nabla \phi(\w_{t+1}) - \nabla \phi (\w_t) \rangle \\
      & = \frac{1}{\eta}  \sum_{t=\I_s}^{\I_{s+1}-1} \langle \bar \e_{\I_s}, \nabla \phi(\w_{t+1}) - \nabla \phi (\w_t) \rangle \\
      & = \frac{1}{\eta} \langle \bar \e_{\I_s}, \nabla \phi(\w_{\I_{s+1}}) - \nabla \phi (\w_{\I_s}) \rangle \\
      & = \frac{1}{\eta} \sum_{i=1}^K  \bar \e_{\I_s}[i] \log{\frac{\w_{\I_{s+1}}[i]}{\w_{\I_s}[i]}} \\
      & \leq \frac{1}{\eta} \sum_{i=1}^K  \bar \e_{\I_s}[i] \log{(KT/S)}  = \frac{\log{(KT/S)}}{\eta} .
   \end{split}
\end{equation}
Combining (\ref{eq:pf-thm1:dec}) with (\ref{eq:pf-thm1:w-t-t1}) and (\ref{eq:pf-thm1-wt1e}) gives
\begin{equation*}
   \begin{split}
      \sum_{t=\I_s}^{\I_{s+1}-1} \langle \w_t - \bar \e_t, \lh_t \rangle & = \sum_{t=\I_s}^{\I_{s+1}-1} \langle \w_t - \w_{t+1}, \lh_t \rangle + \sum_{t=\I_s}^{\I_{s+1}-1} \langle \w_{t+1} - \bar \e_t, \lh_t \rangle \\
      & \leq \sum_{t=\I_s}^{\I_{s+1}-1} \eta + \frac{\log{(KT/S)}}{\eta} \\
      & = \eta (\I_{s+1} - \I_{s}) + \frac{\log{(KT/S)}}{\eta} .
   \end{split}
\end{equation*}
Substituting the above inequality into (\ref{pf-tr-omd-sp-1}), we obtain
\begin{equation*}
   \begin{split}
      \TR(T, S) & \leq \sum_{s=1}^S \sum_{t = \I_s}^{\I_{s+1}-1} \langle \w_t - \bar \e_t, \lh_t \rangle + S \\
      & \leq \sum_{s=1}^S \eta (\I_{s+1} - \I_{s}) + \sum_{s=1}^S \frac{\log{(KT/S)}}{\eta} + S \\
      & = \eta (\I_{S+1} - \I_1) + \frac{S\log{(KT/S)}}{\eta} + S \\
      & = \eta T + \frac{S\log{(KT/S)}}{\eta} + S
   \end{split}
\end{equation*}
which concludes the proof.

\section{Proof of Theorem \ref{thm:PCS}}
\label{sec:pf-pcs}
Following the proof of Theorem \ref{thm:OMD-A} in Appendix \ref{sec:pf-omd-th1}, we have
\begin{equation}
   \label{eq:tr:common}
   \TR(T,S) \leq \frac{1}{\eta} \sum_{s=1}^S \sum_{t = \I_s}^{\I_{s+1}-1} \langle \w_t - \bar \e_t, \eta \lh_t \rangle + S .
\end{equation}
We first decompose $\langle \w_t - \bar  \e_t, \eta \lh_t \rangle$ as
\begin{equation}
   \label{eq:dec-wtetbar-pf-pcs}
   \begin{split}
      & \langle \w_t - \bar  \e_t, \eta \lh_t \rangle \\ 
      = &  \langle \w_t - \w_{t+1}, \eta \lh_t \rangle 
    + \langle \w_{t+1} - \bar  \e_t, \eta \lh_t + \log{(1 - \eta \lh_t)} \rangle
       +  \langle \w_{t+1} - \bar  \e_t, -\log{(1 - \eta \lh_t)} \rangle. 
   \end{split}
\end{equation}
Let $f(\w)$ be a convex function defined as
\begin{equation*}
   f(\w) = \langle \w, -\log{(1 - \eta \lh_t)} \rangle + \D_{\phi} (\w \Vert \w_t), ~\w \in \widetilde \Delta_K .
\end{equation*}
Then, Step 5 of Algorithm \ref{alg:PCS} is identical to
\begin{equation*}
   \w_{t+1} = \mathop{\arg\min}_{\w \in \widetilde \Delta_K} \, f(\w) .
\end{equation*}
By the first order optimal condition, we have
\begin{equation*}
   \langle \w_{t+1} - \u, \nabla f (\w_{t+1}) \rangle \leq 0, ~\forall \u \in \widetilde \Delta_K .
\end{equation*}
Substituting $\u = \bar \e_t$ into the above inequality gives
\begin{equation*}
   \begin{split}
      \langle \w_{t+1} - \bar \e_t, -\log{(1 - \eta \lh_t)} \rangle 
      \leq \langle \bar \e_t - \w_{t+1}, \nabla \phi(\w_{t+1}) - \nabla \phi(\w_t) \rangle
   \end{split}
\end{equation*}
which, together with the decomposition in (\ref{eq:dec-wtetbar-pf-pcs}), leads to
\begin{align}
   & \langle \w_t - \bar  \e_t, \eta \lh_t \rangle \nonumber \\
   \leq \, & \langle \w_t - \w_{t+1}, \eta \lh_t \rangle  + \langle \w_{t+1} - \bar  \e_t, \eta \lh_t + \log{(1 - \eta \lh_t)} \rangle + \langle \bar \e_t - \w_{t+1}, \nabla \phi(\w_{t+1}) - \nabla \phi(\w_t) \rangle \nonumber \\
   = \, & \underbrace{\langle \w_t, \eta \lh_t \rangle + \langle \w_{t+1}, \log{(1 - \eta \lh_t)} - \nabla \phi(\w_{t+1}) +  \nabla \phi(\w_t) \rangle}_{A_t} + \underbrace{\langle -\bar \e_t, \eta \lh_t + \log{(1 - \eta \lh_t)} \rangle}_{B_t} \label{eq:pf-thm-pcs-bd-sp} \\
   &~~~~~~~~~~~~~~~~~~~~~~~~~~~~~~~~~~~~~~~~~~~~~~~~~~~~~~~\quad\quad\quad\quad\quad\,\,\,+ \underbrace{\langle \bar \e_t, \nabla \phi(\w_{t+1}) - \nabla \phi(\w_t) \rangle}_{C_t} . \nonumber
\end{align}
Below, we bound $A_t, B_t$ and  $\sum_{t = \I_s}^{\I_{s+1}-1} C_t$ separately.

(i) Bounding $A_t$.
By the definition of $\phi$ in (\ref{def:neg-entr}), we have
\begin{equation*}
   A_t = \langle \w_t, \eta \lh_t \rangle + \sum_{i=1}^K  \w_{t+1}[i] \log{\frac{(1-\eta\lh_t[i])\w_t[i]}{\w_{t+1}[i]}} .
\end{equation*}
Then, we introduce $\p_{t+1} \in \Delta_K$ defined as
\begin{equation*}
   \p_{t+1}[i] = \frac{\w_{t}[i](1-\eta \lh_t[i])}{\sum_{j=1}^K \w_{t}[j](1-\eta \lh_t[j])}, ~\forall i \in [K]
\end{equation*}
and get
\begin{align}
      A_t = \, & \langle \w_t, \eta \lh_t \rangle + \sum_{i=1}^K  \w_{t+1}[i] \log{\frac{(1-\eta\lh_t[i])\w_t[i]}{\w_{t+1}[i]}} \nonumber \\ 
      + & \sum_{i=1}^K  \w_{t+1}[i] \log{\frac{\w_{t+1}[i]}{\p_{t+1}[i]}} 
      - \D_{\phi}(\w_{t+1} \Vert \p_{t+1}) \nonumber \\
      \leq \, & \langle \w_t, \eta \lh_t \rangle + \sum_{i=1}^K \w_{t+1}[i] \log{\frac{(1-\eta\lh_t[i])\w_t[i]}{\p_{t+1}[i]}}  \nonumber \\
      = \, & \langle \w_t, \eta \lh_t \rangle + \sum_{i=1}^K  \w_{t+1}[i] \log{\sum_{j=1}^K \w_{t}[j](1-\eta \lh_t[j])} \nonumber \\
      = \, & \langle \w_t, \eta \lh_t \rangle + \log{\sum_{j=1}^K \w_{t}[j](1-\eta \lh_t[j])}  \label{eq:pf-thm-pcs-bd-a} \\
      = \, & \langle \w_t, \eta \lh_t \rangle + \log{\left( \sum_{j=1}^K \w_{t}[j] - \sum_{j=1}^K \eta \w_t[j] \lh_t[j] \right)} \nonumber \\
      = \, & \langle \w_t, \eta \lh_t \rangle + \log{(1 -\langle \w_t, \eta \lh_t \rangle)} \nonumber \\
      \leq \, & 0 \nonumber
\end{align}
where the first equality follows from the definition of Bregman divergence and the second equality is due to the definition of $\p_{t+1}$; the first inequality holds since Bregman divergence is always non-negative, and the last inequality holds since $\langle \w_t, \eta \lh_t \rangle \in [0, 1/2]$ and $x + \log{(1-x)} \leq 0, ~\forall x \in [0, 1) $.

(ii) Bounding $B_t$. By the fact that $\forall t \in [T], \eta\lh_t \in [0,1/2]^K$ and the well-known inequality $, \forall x \in (-\infty, 1/2], -x - \log{(1-x)} \leq x^2$, we have
\begin{align}
      B_t & = \sum_{i=1}^K \bar \e_t[i] \big( - \eta \lh_t[i] - \log{(1-\eta \lh_t[i])} \big) \nonumber \\
      & \leq \sum_{i=1}^K \bar \e_t[i] (\eta \lh_t[i])^2  = \sum_{i=1}^K \left(1-\frac{S}{T}\right) \e_t[i] (\eta \lh_t[i])^2  + \sum_{i=1}^K \frac{S (\eta \lh_t[i])^2}{TK} \nonumber \\
      & \leq \sum_{i=1}^K \e_t[i] (\eta \lh_t[i])^2 + \sum_{i=1}^K  \frac{\eta S}{2TK} \label{eq:pf-thm-pcs-bd-b} \\
      & = \eta^2 (\lh_t[\EP_t^*])^2 + \frac{\eta S}{2T} . \nonumber
\end{align}

(iii) Bounding $\sum_{t = \I_s}^{\I_{s+1}-1} C_t$. Following (\ref{eq:pf-thm1-wt1e}), we have
\begin{equation}
   \label{eq:pf-thm-pcs-bd-c}
   \sum_{t = \I_s}^{\I_{s+1}-1} C_t \leq \log{(KT/S)} . 
\end{equation}

Combining (\ref{eq:tr:common})--(\ref{eq:pf-thm-pcs-bd-c}), we have
\begin{equation*}
   \begin{split}
      \TR(T,S) & \leq \frac{1}{\eta} \sum_{s=1}^S \sum_{t = \I_s}^{\I_{s+1}-1} \left(\eta^2 (\lh_t[\EP_t^*])^2 + \frac{\eta S}{2T}\right) + \frac{1}{\eta} \sum_{s=1}^S \log{(KT/S)} + S \\
      & = \eta \sum_{t=1}^T (\lh_t[\EP_t^*])^2 + \frac{S \log{(KT/S)}}{\eta} + \frac{3S}{2}
   \end{split}
\end{equation*}
which finishes the proof.

\section{Proof of Theorem \ref{thm:OCS}}
\label{app:pf-thm3}
Following the proof of Theorem \ref{thm:OMD-A} in Appendix \ref{sec:pf-omd-th1}, we have
\begin{equation}
   \label{pf-tr-co-3}
      \TR(T,S) \leq \sum_{s=1}^S \sum_{t = \I_s}^{\I_{s+1}-1} \langle \w_t - \bar \e_t, \lh_t \rangle + S .
\end{equation}
We start by splitting $\langle \w_t - \bar \e_t, \lh_t \rangle$ into three terms:
\begin{equation}
   \label{eq:split-w-bare}
   \begin{split}
      \langle \w_t - \bar  \e_t, \lh_t \rangle & = \langle \w_t - \widetilde \w_{t+1}, \lh_t \rangle + \langle \widetilde \w_{t+1} - \bar  \e_t, \lh_t \rangle \\
      & = \langle \w_t - \widetilde \w_{t+1}, \lh_t - \lh_{t-1} \rangle + \langle \w_t - \widetilde \w_{t+1}, \lh_{t-1} \rangle + \langle \widetilde \w_{t+1} - \bar  \e_t, \ell_t \rangle .
   \end{split}
\end{equation}
The first term can be bounded by the following lemma.
\begin{lma}
   \label{lm:w-t-t1}
   For any $t \in [T]$, we have
   \begin{equation}
      \label{eq:w-t-t1}
      \langle \w_t - \widetilde \w_{t+1}, \lh_t - \lh_{t-1} \rangle \leq \eta \Vert \lh_t - \lh_{t-1} \Vert_{\infty}^2 .
   \end{equation}
\end{lma}
To bound the second and the third terms, we define two convex functions on the clipped simplex $\widetilde \Delta_K$:
\begin{equation*}
   \begin{split}
      f(\w) & = \langle \w, \eta \lh_{t-1} \rangle + \D_{\phi}(\w \Vert \widetilde \w_{t}), ~\w \in  \widetilde \Delta_K; \\
      g(\w) & = \langle \w, \eta \lh_t \rangle + \D_{\phi}(\w \Vert \widetilde \w_t), ~\w \in \widetilde \Delta_K.
   \end{split}
\end{equation*}
Then, we can rewrite Steps $3$ and $5$ in Algorithm \ref{alg:OCS} as
\begin{equation*}
      \w_{t} = \mathop{\arg\min}_{\w \in \widetilde \Delta_K} f(\w) ; ~~~~~~
      \widetilde \w_{t+1} = \mathop{\arg\min}_{\w \in \widetilde \Delta_K} g(\w) .
\end{equation*}
By the first order optimal condition, we have
\begin{equation*}
   \langle \w_t - \u, \nabla f(\w_t) \rangle \leq 0, ~\forall \u \in \widetilde \Delta_K; ~~~~~~ \langle \widetilde \w_{t+1} - \v, \nabla g(\widetilde \w_{t+1}) \rangle \leq 0, ~\forall \v \in \widetilde \Delta_K.
\end{equation*}
Substituting $\u = \widetilde \w_{t+1}$ and $\v = \bar \e_t$ into the above two inequalities respectively, we get
\begin{equation}
   \label{eq:u-w}
   \begin{split}
      \langle \w_t - \widetilde \w_{t+1}, \nabla f(\w_t) \rangle & \leq 0 \\
      \langle \w_t - \widetilde \w_{t+1},  \eta \lh_{t-1} + \nabla \phi(\w_t) -  \nabla \phi(\widetilde \w_t) \rangle & \leq 0 \\
      \langle \w_t - \widetilde \w_{t+1}, \lh_{t-1} \rangle & \leq \frac{1}{\eta} \langle \w_t - \widetilde \w_{t+1}, \nabla \phi(\widetilde \w_t) -  \nabla \phi(\w_t) \rangle ;
   \end{split}
\end{equation}
and
\begin{equation}
   \label{eq:v-e}
   \begin{split}
      \langle \widetilde \w_{t+1} - \bar \e_t, \nabla g(\widetilde \w_{t+1}) \rangle & \leq 0 \\
      \langle \widetilde \w_{t+1} - \bar \e_t,  \eta \lh_t + \nabla \phi(\widetilde \w_{t+1}) -  \nabla \phi(\widetilde \w_t) \rangle & \leq 0 \\
      \langle \widetilde \w_{t+1} - \bar \e_t,  \lh_t \rangle & \leq \frac{1}{\eta} \langle \widetilde \w_{t+1} - \bar \e_t, \nabla \phi(\widetilde \w_t) -  \nabla \phi(\widetilde \w_{t+1}) \rangle .
   \end{split} 
\end{equation}
Combining (\ref{eq:u-w}) and (\ref{eq:v-e}) and rearranging, we have
\begin{equation*}
   \begin{split}
   & ~ \langle \w_t - \widetilde \w_{t+1}, \lh_{t-1} \rangle + \langle \widetilde \w_{t+1} - \bar \e_t,  \lh_t \rangle \\ 
   \leq & ~ \frac{1}{\eta} \Big( \langle \w_t, \nabla \phi(\widetilde \w_t) -  \nabla \phi(\w_t) \rangle - \langle \widetilde \w_{t+1}, \nabla \phi(\widetilde \w_t) -  \nabla \phi(\w_t) \rangle  \\ 
   & + \langle \widetilde \w_{t+1}, \nabla \phi(\widetilde \w_t) -  \nabla \phi(\widetilde \w_{t+1}) \rangle + \langle \bar \e_t, \nabla \phi(\widetilde \w_{t+1}) -  \nabla \phi(\widetilde \w_t) \rangle \Big) \\
   = & ~ \frac{1}{\eta} \Big( \langle \w_t, \nabla \phi(\widetilde \w_t) -  \nabla \phi(\w_t) \rangle + \langle \widetilde \w_{t+1}, \nabla \phi(\w_t) -  \nabla \phi(\widetilde \w_{t+1}) \rangle + \langle \bar \e_t, \nabla \phi(\widetilde \w_{t+1}) -  \nabla \phi(\widetilde \w_t) \rangle \Big) \\
   = & ~ \frac{1}{\eta} \Big( -\D_\phi(\w_t \Vert \widetilde \w_t) - \D_\phi(\widetilde \w_{t+1} \Vert \w_t) + \langle \bar \e_t, \nabla \phi(\widetilde \w_{t+1}) -  \nabla \phi(\widetilde \w_t) \rangle \Big) \\
   \leq & ~ \frac{1}{\eta} \langle \bar \e_t, \nabla \phi(\widetilde \w_{t+1}) -  \nabla \phi(\widetilde \w_t) \rangle
   \end{split}
\end{equation*}
where the last equality follows from the definition of Bregman divergence in (\ref{def:bregman}), and the last inequality holds since Bregman divergence is always non-negative.
Substituting the above inequality and (\ref{eq:w-t-t1}) into (\ref{eq:split-w-bare}), we get
\begin{equation*}
   \langle \w_t - \bar  \e_t, \lh_t \rangle \leq \eta \Vert \lh_t - \lh_{t-1} \Vert_{\infty}^2 + \frac{1}{\eta} \langle \bar \e_t, \nabla \phi(\widetilde \w_{t+1}) -  \nabla \phi(\widetilde \w_t) \rangle .
\end{equation*}
Summing this inequality over $t=\I_s, \ldots \I_{s+1}-1$ and following the same derivation as in (\ref{eq:pf-thm1-wt1e}), we have
\begin{equation*}
   \begin{split}
      \sum_{t = \I_s}^{\I_{s+1}-1} \langle \w_t - \bar \e_t, \lh_t \rangle 
      & \leq \sum_{t = \I_s}^{\I_{s+1}-1} \eta \Vert \lh_t - \lh_{t-1} \Vert_{\infty}^2 + \frac{1}{\eta} \sum_{t = \I_s}^{\I_{s+1}-1} \langle \bar \e_t, \nabla \phi(\widetilde \w_{t+1}) -  \nabla \phi(\widetilde \w_t) \rangle \\
      & \leq \sum_{t = \I_s}^{\I_{s+1}-1} \eta \Vert \lh_t - \lh_{t-1} \Vert_{\infty}^2 + \frac{\log{(KT/S)}}{\eta} .
   \end{split}
\end{equation*}
Substituting the above inequality into (\ref{pf-tr-co-3}), we get
\begin{equation*}
   \begin{split}
      \TR(T, S) & \leq \sum_{s=1}^S \sum_{t = \I_s}^{\I_{s+1}-1} \langle \w_t - \bar \e_t, \lh_t \rangle + S \\
      & \leq \sum_{s=1}^S \left( \sum_{t = \I_s}^{\I_{s+1}-1} \eta \Vert \lh_t - \lh_{t-1} \Vert_{\infty}^2 + \frac{\log{(KT/S)}}{\eta} \right) + S \\
      & = \eta P_{\infty} + \frac{S \log{(KT/S)}}{\eta} + S .
   \end{split}
\end{equation*}
This completes the proof.

\section{Proof of Theorem \ref{thm:OCSD}}
Let $m^*$ be the last epoch such that
\begin{equation*}
   m^* = \max ~\{ m : \tau_m < T \}
\end{equation*}
and define $\tau_{m^*+1} = T$. We begin with bounding the tracking regret in each epoch $m=1,\ldots,m^*$. Specifically, considering the $m$-th epoch, by the proof of Theorem \ref{thm:OCS} in Appendix \ref{app:pf-thm3}, we have
\begin{equation}
   \label{eq:tr-m-epoch}
   \begin{split}
      & \sum_{t=\tau_m+1}^{\tau_{m+1}} \hat \ell_t - \sum_{t=\tau_m+1}^{\tau_{m+1}} \lh_t[\EP_t^*] \\ 
      \leq \, & \eta_m P_m + \frac{S\log{(KT/S)}}{\eta_m} + \frac{S(\tau_{m+1} - \tau_m)}{T} \\
      = \, & \eta_m \sum_{t=\tau_m+1}^{\tau_{m+1}} \Vert \lh_t - \lh_{t-1} \Vert_\infty^2 + \frac{S\log{(KT/S)}}{\eta_m} + \frac{S(\tau_{m+1} - \tau_m)}{T} \\
      \leq \, & \eta_m \sum_{t=\tau_m+1}^{\tau_{m+1} - 1} \Vert \lh_t - \lh_{t-1} \Vert_\infty^2 + \eta_m  + \frac{S\log{(KT/S)}}{\eta_m} + \frac{S(\tau_{m+1} - \tau_m)}{T} \\
      \leq \, & \frac{S\log{(KT/S)}}{\eta_m} + \eta_m + \frac{S\log{(KT/S)}}{\eta_m} + \frac{S(\tau_{m+1} - \tau_m)}{T} \\
      = \, & \frac{2S\log{(KT/S)}}{\eta_m} + \eta_m + \frac{S(\tau_{m+1} - \tau_m)}{T}
   \end{split}
\end{equation}
where the second inequality is due to the fact that $\lh_t \in [0, 1]^K, ~\forall t \in [T]$, and the last inequality holds since for each epoch, the condition in Line $10$ of Algorithm \ref{alg:OCSPlus} can be violated only at the last round of the epoch.
Summing (\ref{eq:tr-m-epoch}) over $m=1,\ldots,m^*$, we get
\begin{equation*}
\begin{split}
   \TR(T, S) & = \sum_{t=1}^T \widehat \ell_t - \sum_{t=1}^T \lh_t[\EP_t^*] = \sum_{m=1}^{m^*} \sum_{t=\tau_m+1}^{\tau_{m+1}} \widehat \ell_t - \sum_{m=1}^{m^*} \sum_{t=\tau_m+1}^{\tau_{m+1}} \lh_t[\EP_t^*] \\
   & \leq \sum_{m=1}^{m^*} \left( \frac{2S\log{(KT/S)}}{\eta_m} + \eta_m + \frac{S(\tau_{m+1} - \tau_m)}{T} \right) \\
   & = \sum_{m=1}^{m^*} \frac{2S\log{(KT/S)}}{\eta_m} + \sum_{m=1}^{m^*} \eta_m + \frac{S(\tau_{{m^*}+1} - \tau_1)}{T} \\ 
   & = \sum_{m=1}^{m^*} \frac{2S\log{(KT/S)}}{\eta_m} + \sum_{m=1}^{m^*} \eta_m + S .
\end{split}
\end{equation*}
By the update rule of $\eta_m$ (Line $11$ in Algorithm \ref{alg:OCSPlus}), we have $\eta_m = \frac{\sqrt{S\log{(KT/S)}}}{2^{m-1}}$ and thus
\begin{equation}
   \label{eq:tr-OCSD}
   \begin{split}
      \TR(T, S) & \leq \sqrt{S\log{(KT/S)}} \sum_{m=1}^{m^*} 2^m  + \sqrt{S\log{(KT/S)}} \sum_{m=1}^{m^*} \frac{1}{2^{m-1}} + S \\
      & \leq (2^{m^*+1} - 2) \sqrt{S\log{(KT/S)}} + 2 \sqrt{S\log{(KT/S)}} + S \\
      & = 2^{m^*+1} \sqrt{S\log{(KT/S)}} + S .
   \end{split}
\end{equation}
Below we consider two cases:

(i) $m^* = 1$. In this case, it trivially follows that
\begin{equation}
   \label{eq:pf-OCSD-case1}
   \TR(T, S) \leq 4\sqrt{S\log{(KT/S)}} + S \leq \O(\sqrt{S (P_\infty+1) \log{(KT/S)}} + S).
\end{equation}

(ii) $m^* > 1$. In this case, since the $(m^*-1)$-th epoch has finished, we have
\begin{equation*}
   \begin{split}
      \eta_{m^* - 1} >  \sqrt{\frac{S\log{(KT/S)}}{P_{m^*-1}}}
   \end{split}
\end{equation*}
which implies
\begin{equation*}
   \begin{split}
      \frac{\sqrt{S\log{(KT/S)}}}{2^{m^*-2}} & > \sqrt{\frac{S\log{(KT/S)}}{P_{m^*-1}}} \\
      2^{m^*-2} & < \sqrt{P_{m^*-1}} \\
      2^{m^*+1} & < 8 \sqrt{P_{m^*-1}}.
   \end{split}
\end{equation*}
Substituting the above inequality into (\ref{eq:tr-OCSD}) gives
\begin{equation}
   \label{eq:pf-OCSD-case2}
   \begin{split}
      \TR(T, S) & \leq 8 \sqrt{P_{m^*-1}S\log{(KT/S)}} + S \\
      & \leq 8 \sqrt{P_\infty S\log{(KT/S)}} + S \\
      & \leq \O(\sqrt{S(P_\infty+1)\log{(KT/S)}} + S) .
   \end{split}
\end{equation}

Combining (\ref{eq:pf-OCSD-case1}) and (\ref{eq:pf-OCSD-case2}) completes the proof.

\section{Proof of Theorem \ref{thm:PCSP}}
The proof below is a generalization of the proof of Theorem \ref{thm:PCS} in Appendix \ref{sec:pf-pcs}. Similarly to (\ref{eq:samee}), we first divide the time horizon $[1, T]$ into $S$ disjoint intervals $[\I_1, \I_2), \ldots, [\I_S, \I_{S+1})$ with $\I_1 = 1$ and $\I_{S+1} = T+1$ such that in each interval $[\I_s, \I_{s+1}), s \in [S]$, the compared matrix $U_t^*$ remains the same, i.e., 
\begin{equation}
   \label{eq:samee-mtx}
   U_{\I_s}^* = U_{\I_s+1}^* = U_{\I_s+2}^* =  \cdots = U_{\I_{s+1} - 1}^*, ~\forall s \in [S] .
\end{equation}
Fix $s \in [S]$. We consider the tracking regret in the $s$-th interval:
\begin{equation}
   \label{eq:tr-s-mtx}
   \sum_{t=\I_s}^{\I_{s+1}-1} \tr(W_t Z_t) - \sum_{t=\I_s}^{\I_{s+1}-1} \tr(U_t^* Z_t) = \sum_{t=\I_s}^{\I_{s+1}-1} \tr\big((W_t - U_t^*) Z_t\big) .
\end{equation}
Let $\bar U_t^*$ be defined by
\begin{equation}
   \label{eq:def:barut}
   \bar U_t^* = (1-\frac{S}{T}) U_t^* + \frac{SI_K}{TK} = (1-\frac{S}{T}) U_t^* + \frac{SI}{TK}
\end{equation}
in which (and in the following) the subscript $K$ of the $K\times K$ identity matrix $I_K$ is omitted for brevity.
We decompose the right-hand side of (\ref{eq:tr-s-mtx}) as
\begin{equation}
   \label{eq:tr-s-1-mtx}
   \sum_{t=\I_s}^{\I_{s+1}-1} \tr\big((W_t - U_t^*) Z_t\big) = \sum_{t=\I_s}^{\I_{s+1}-1} \tr\big((W_t - \bar U_t^*) Z_t\big) + \sum_{t=\I_s}^{\I_{s+1}-1} \tr\big((\bar U_t^* - U_t^*) Z_t\big)
\end{equation}
where $\tr\big((\bar U_t^* - U_t^*)Z_t \big) $ can be bounded by the following lemma.
\begin{lma}
   \label{lm:mtx-baru-u}
   For any $t \in [T]$, we have
   \begin{equation}
      \label{eq:tr-s-2-mtx}
      \tr\big((\bar U_t^* - U_t^*) Z_t\big) \leq \frac{2S}{T} .
   \end{equation}
\end{lma}
Below we focus on bounding $\eta \tr\big((W_t - \bar U_t^*) Z_t\big) = \tr\big((W_t - \bar U_t^*) (\eta Z_t) \big)$ and start by  splitting it into three terms:
\begin{equation}
   \label{eq:split-mtx}
   \begin{split}
      & \tr\big((W_t - \bar U_t^*) (\eta Z_t) \big) \\
      = \,& \tr\big( (W_t - W_{t+1}) (\eta Z_t) \big)  + \tr\big( (W_{t+1} - \bar  U_t^*) (\eta Z_t + \log{(I - \eta Z_t)}) \big) \\
      & + \tr\big( (W_{t+1} - \bar  U_t^*) (-\log{(I - \eta Z_t)}) \big) .
   \end{split}
\end{equation}
Then, we introduce a convex function on the clipped spectraplex $\widetilde \Omega_K$:
\begin{equation*}
   H(W) = \tr(-W\log{(I - \eta Z_t)}) + \D_{\psi} (W \Vert W_t), ~ W \in \widetilde \Omega_K
\end{equation*}
and rewrite Step $5$ of Algorithm \ref{alg:PCSP} as
\begin{equation*}
   W_{t+1} = \mathop{\arg\min}_{W \in \widetilde \Omega_K} \, H(W) .
\end{equation*}
By the first order optimal condition and the fact that $\bar U_t^* \in \widetilde \Omega_K$, we have
\begin{equation*}
   \tr\left((W_{t+1} - \bar U_t^*)\nabla H(W_{t+1})\right) \leq 0 .
\end{equation*}
Expanding $\nabla H(W_{t+1})$ and using the equality $\nabla \psi(W) = I + \log{W}$, we get
\begin{equation*}
   \tr\Big(\big(W_{t+1} - \bar U_t^* \big)\big(-\log{(I - \eta Z_t)} + \log{W_{t+1}} - \log{W_t}\big)\Big) \leq 0
\end{equation*}
which implies
\begin{equation*}
   \tr\Big( \big(W_{t+1} - \bar U_t^* \big)\big(-\log{(I - \eta Z_t)}\big) \Big) \leq \tr\Big(\big(\bar U_t^* - W_{t+1}\big)\big(\log{W_{t+1}} - \log{W_t}\big)\Big) .
\end{equation*}
Combining the above inequality with (\ref{eq:split-mtx}) gives
\begin{equation*}
   \begin{split}
      & \tr\big((W_t - \bar U_t^*) (\eta Z_t) \big) \\
      \leq \,& \tr\big( (W_t - W_{t+1}) (\eta Z_t) \big)  + \tr\big( (W_{t+1} - \bar  U_t^*) (\eta Z_t + \log{(I - \eta Z_t)}) \big) \\
      & + \tr\big((\bar U_t^* - W_{t+1})(\log{W_{t+1}} - \log{W_t})\big) \\
      = \, & \underbrace{\tr\big( (W_t - W_{t+1}) (\eta Z_t) \big)  + \tr \big(W_{t+1} (\eta Z_t + \log{(I - \eta Z_t)} - \log{W_{t+1}} +  \log{W_t}) \big)}_{A_t} \\
      & + \underbrace{\tr \big(-\bar U_t^*(\eta Z_t + \log{(I - \eta Z_t)}) \big)}_{B_t} + \tr \big( \bar U_t^*(\log{W_{t+1}} -  \log{W_t}) \big) .
   \end{split}
\end{equation*}
The following lemmas bound $A_t$ and $B_t$ respectively:
\begin{lma}
   \label{lm:PCSP-1}
   For any $t \in [T]$, we have
   \begin{equation*}
      A_t = \tr\big( (W_t - W_{t+1}) (\eta Z_t) \big)  + \tr \big(W_{t+1} (\eta Z_t + \log{(I - \eta Z_t)} - \log{W_{t+1}} +  \log{W_t}) \big) \leq 0 .
   \end{equation*}
\end{lma}
\begin{lma}
   \label{lm:PCSP-2}
   For any $t \in [T]$, we have
   \begin{equation*}
      B_t = \tr \big(-\bar U_t^*(\eta Z_t + \log{(I - \eta Z_t)}) \big) \leq \eta^2 \tr\big(U_t^* Z_t^2 \big) + \frac{\eta S}{2T} .
   \end{equation*}
\end{lma}
It follows that
\begin{equation}
   \label{eq:eta-tr-mtx}
   \begin{split}
      & \sum_{t=\I_s}^{\I_{s+1}-1} \tr\big((W_t - \bar U_t^*) (\eta Z_t)\big) \\
      \leq \, & \, 0 + \sum_{t=\I_s}^{\I_{s+1}-1} \eta^2 \tr\big(U_t^* Z_t^2 \big) + \sum_{t=\I_s}^{\I_{s+1}-1} \frac{\eta S}{2T} + \sum_{t=\I_s}^{\I_{s+1}-1} \tr \big( \bar U_t^*(\log{W_{t+1}} -  \log{W_t}) \big) \\
      = \, & \sum_{t=\I_s}^{\I_{s+1}-1} \eta^2 \tr\big(U_t^* Z_t^2 \big) + \frac{\eta S(\I_{s+1} - \I_s)}{2T} + \sum_{t=\I_s}^{\I_{s+1}-1} \tr \big( \bar U_{\I_s}^* (\log{W_{t+1}} -  \log{W_t}) \big) \\
      = \, & \sum_{t=\I_s}^{\I_{s+1}-1} \eta^2 \tr\big(U_t^* Z_t^2 \big) + \frac{\eta S(\I_{s+1} - \I_s)}{2T} + \tr \big( \bar U_{\I_s}^* (\log{W_{\I_{s+1}}} -  \log{W_{\I_s}}) \big) \\
      \leq \, & \sum_{t=\I_s}^{\I_{s+1}-1} \eta^2 \tr\big(U_t^* Z_t^2 \big) + \frac{\eta S(\I_{s+1} - \I_s)}{2T} + \log{(KT/S)}
   \end{split}
\end{equation}
where the first equality holds since $\bar U_{\I_s}^* = \bar U_{\I_s+1}^* = \bar U_{\I_s+2}^* =  \cdots = \bar U_{\I_{s+1} - 1}^*$, and the second inequality is due do the following lemma.
\begin{lma}
   \label{lm:logkt-mtx}
   For any $X, Y, Z \in \widetilde \Omega_K$, we have
   \begin{equation*}
      \tr \big(X (\log{Y} - \log{Z}) \big) \leq \log{(KT/S)}.
   \end{equation*}
\end{lma}
Dividing both sides of (\ref{eq:eta-tr-mtx}) by $\eta$ and summing over $s=1,\ldots,S$ leads to
\begin{equation*}
   \begin{split}
      \sum_{s=1}^S \sum_{t=\I_s}^{\I_{s+1}-1} \tr\big((W_t - \bar U_t^*) Z_t\big) & \leq \sum_{s=1}^S \sum_{t=\I_s}^{\I_{s+1}-1} \eta  \tr\big(U_t^* Z_t^2 \big) + \frac{S(\I_{S+1} - \I_1)}{2T} + \frac{S\log{(KT/S)}}{\eta} \\
      & = \eta \sum_{t=1}^T \tr\big(U_t^* Z_t^2 \big) + \frac{S}{2} + \frac{S\log{(KT/S)}}{\eta} .
   \end{split}
\end{equation*}
Combining the above inequality with (\ref{eq:tr-s-1-mtx}) and (\ref{eq:tr-s-2-mtx}), we have
\begin{equation*}
   \begin{split}
      \sum_{s=1}^S \sum_{t=\I_s}^{\I_{s+1}-1} \tr\big((W_t - U_t^*) Z_t\big) & = \sum_{s=1}^S \sum_{t=\I_s}^{\I_{s+1}-1} \tr\big((W_t - \bar U_t^*) Z_t\big) + \sum_{s=1}^S \sum_{t=\I_s}^{\I_{s+1}-1} \tr\big((\bar U_t^* - U_t^*) Z_t\big) \\
      & \leq \eta \sum_{t=1}^T \tr\big(U_t^* Z_t^2 \big) + \frac{S}{2} + \frac{S\log{(KT/S)}}{\eta} + \sum_{s=1}^S \sum_{t=\I_s}^{\I_{s+1}-1} \frac{2S}{T} \\
      & = \eta M_2 + \frac{S\log{(KT/S)}}{\eta} + \frac{5S}{2} .
   \end{split}
\end{equation*}

\section{Proofs of Lemmas}
In this appendix, we provide the proofs of all lemmas.

\subsection{Proof of Lemma \ref{lm-prop-1}}
Define 
\begin{equation*}
   C = \sum_{j=1}^K  \widehat \w_t[j] \exp(-\eta \lh_t[j]) .
\end{equation*}
We can rewrite (\ref{eq:pf-prop-up0}) as
\begin{equation*}
   \widehat \w_{t+1}^m[i]  = \frac{\widehat \w_t[i] \exp(-\eta \lh_t[i])}{C}, \, \forall i \in [K] .
\end{equation*}
By the definition of $\phi$ in (\ref{def:neg-entr}), for any $i \in [K]$ we have
\begin{equation*}
   \eta \lh_t[i] +  \nabla \phi(\widehat \w_{t+1}^m)[i] - \nabla \phi (\widehat \w_t)[i] = \eta \lh_t[i] + \log{\left( \frac{\widehat \w_t[i] \exp(-\eta \lh_t[i])}{C} \right)} - \log{(\widehat \w_t [i]) } = - \log{C} .
\end{equation*}
It follows that
\begin{equation*}
   \begin{split}
      & \langle \u - \widehat \w_{t+1}^m , \eta \lh_t + \nabla \phi(\widehat \w_{t+1}^m) - \nabla \phi (\widehat \w_t) \rangle \\
      = \, & \sum_{i=1}^K \big(\u[i] - \widehat \w_{t+1}^m[i]\big) \big(\eta \lh_t[i] + \nabla \phi(\widehat \w_{t+1}^m)[i] - \nabla \phi (\widehat \w_t)[i]\big) \\
      = \, & -\big(\log{C}\big) \sum_{i=1}^K \big(\u[i] - \widehat \w_{t+1}^m[i] \big) = -\big(\log{C}\big) \left( \sum_{i=1}^K \u[i] - \sum_{i=1}^K \widehat \w_{t+1}^m[i] \right) = 0
   \end{split}
\end{equation*}
where the last inequality holds since $\sum_{i=1}^K \u[i] = 1$ and $\sum_{i=1}^K \widehat \w_{t+1}^m[i] = 1$.

\subsection{Proof of Lemma \ref{lm:bare-e-lt}}
By the definition of $\bar \e_t$ in ({\ref{def:bar-e}}), we have
\begin{equation*}
   \bar \e_t[i] - \e_t[i] = (1-\frac{S}{T}) \e_t[i] + \frac{S}{TK} - \e_t[i] = -\frac{S\e_t[i]}{T} + \frac{S}{TK}, ~\forall i \in [K] .
\end{equation*}
It follows that
\begin{equation*}
   \begin{split}
      \langle \bar \e_t - \e_t, \lh_t \rangle = \sum_{i=1}^K (\bar \e_t[i] - \e_t[i])\lh_t[i]  & = \sum_{i=1}^K \left(-\frac{S \e_t[i]}{T} + \frac{S}{TK}\right)\lh_t[i]  \\
      & = - \sum_{i=1}^K \frac{S \e_t[i]\lh_t[i]}{T} + \frac{S}{TK} \sum_{i=1}^K \lh_t[i] \\
      & \leq 0 + \frac{S}{TK} \cdot K = \frac{S}{T}
   \end{split}
\end{equation*}
where the inequality holds since $0 \leq \e_t[i], \lh_t[i] \leq 1, ~\forall i \in [K]$.

\subsection{Proof of Lemma \ref{lm:pf-thm1:w-t-t1}}
\label{app:lm:pf-thm1:w-t-t1}
We first introduce the definition of Fenchel conjugate:
\begin{defi}
   Let $\X \subseteq  \mathbb{R}^n $ be a convex set and $f : \X \mapsto \mathbb{R}$ be a convex function. The Fenchel conjugate of $f$ is a function $f^* : \mathbb{R}^n \mapsto \mathbb{R}$, defined as
   \begin{equation*}
      f^*(\y) = \sup_{\x \in \X} ~ \langle \x, \y \rangle - f(\x), ~~~ \y \in \mathbb{R}^n.
   \end{equation*}
\end{defi}
As a powerful tool in convex analysis, the Fenchel conjugate has many properties among which we mainly utilize the following three properties, the proof of which can be found at, e.g., \citet{Shai:thesis}.
\begin{thm}
   \label{thm:duality}
   Let $\X \subseteq  \mathbb{R}^n $ be a convex set and $f : \X \mapsto \mathbb{R}$ be a convex function. If $f$ is further closed and $\mu$-strongly convex with respect to a norm $\Vert \cdot \Vert$, then its Fenchel conjugate function $f^*$ is everywhere differentiable and the gradient of $f^*$ satisfies
   \begin{itemize}
      \item for any $\y \in \mathbb{R}^n$,
      \begin{equation}
      \label{eq:prop1-conj}
      \nabla f^*(\y) = \mathop{\arg\max}_{\x \in \X} ~ \langle \x, \y \rangle - f(\x) ;
      \end{equation}
      \item for any $\y, \z \in \mathbb{R}^n$,
      \begin{equation}
      \label{eq:prop2-conj}
      \Vert \nabla f^*(\y) - \nabla f^*(\z) \Vert \leq \frac{1}{\mu} \Vert \y - \z \Vert_{*}
      \end{equation}
      where $\Vert \cdot \Vert_*$ is the dual norm of $\Vert \cdot \Vert ;$
      \item for any $\x \in \X$,
      \begin{equation}
      \label{eq:prop3-conj}
      \nabla f^*(\nabla f(\x)) = \x.
      \end{equation}
   \end{itemize}
\end{thm}
Let $\widehat \phi$ be the negative Shannon entropy function $\phi$ with domain being the clipped simplex $\widetilde \Delta_K$. It is easy to see that $\widehat \phi$ is closed as $\phi$ is a continuous function and $\widetilde \Delta_K$ is a closed set. Furthermore, it is well-known that $\phi$ and hence $\widehat \phi$ are $1$-strongly convex with respect to the $\Vert \cdot \Vert_1$ norm \citep{Shai:thesis}. Therefore, $\widehat \phi$ meets the condition of Theorem \ref{thm:duality} and $\nabla \widehat \phi^*$ enjoys the three above properties, which play a key role in the analysis below.

Fix $t \in [T]$. By the updating step in (\ref{eq:update-calA}), we have
\begin{equation}
   \label{eq:model-der}
   \begin{split}
      \w_{t+1} & = \mathop{\arg\min}_{\w \in \widetilde \Delta_K} \langle \w, \eta \lh_t \rangle +  \D_{\phi}(\w \Vert \w_t) \\
      & = \mathop{\arg\min}_{\w \in \widetilde \Delta_K} \langle \w, \eta \lh_t \rangle + \phi(\w) - \langle \w, \nabla \phi(\w_t) \rangle \\
      & = \mathop{\arg\max}_{\w \in \widetilde \Delta_K} ~ \langle \w, \nabla \phi(\w_t) - \eta \lh_t \rangle - \phi(\w) \\
      & = \mathop{\arg\max}_{\w \in \widetilde \Delta_K} ~ \langle \w, \nabla \phi(\w_t) - \eta \lh_t \rangle - \widehat \phi(\w) \\
      & = \nabla \widehat \phi^* (\nabla \phi(\w_t) - \eta \lh_t)
   \end{split}
\end{equation}
where the last equality follows from (\ref{eq:prop1-conj}). 
On the other hand, by (\ref{eq:prop3-conj}) we can rewrite $\w_t$ as
\begin{equation*}
   \w_t = \nabla \widehat \phi^* \big(\nabla \widehat \phi(\w_t) \big) = \nabla \widehat \phi^* \big(\nabla \phi(\w_t) \big) .
\end{equation*}
Combining the above two equalities, we get
\begin{equation*}
   \begin{split}
      \langle \w_t - \w_{t+1}, \lh_t \rangle & = \langle \nabla \widehat \phi^* \big(\nabla \phi(\w_t) \big) - \nabla \widehat \phi^* \big(\nabla \phi(\w_t) - \eta \lh_t \big), \lh_t \rangle \\
      & \leq \Vert \nabla \widehat \phi^* \big(\nabla \phi(\w_t)\big) - \nabla \widehat \phi^* \big(\nabla \phi(\w_t) - \eta \lh_t \big) \Vert_1 \Vert \lh_t \Vert_{\infty} \\
      & \leq \Vert \big(\nabla \phi(\w_t) \big) - \big(\nabla \phi(\w_t) - \eta \lh_t \big) \Vert_{\infty} \Vert \lh_t \Vert_{\infty} \\
      & = \eta \Vert \lh_t \Vert_{\infty}^2 \\
      & \leq \eta
   \end{split}
\end{equation*}
where the first inequality follows from the Cauchy–Schwarz inequality, the second inequality is due to ($\ref{eq:prop2-conj}$) and the fact that the dual norm of $\Vert \cdot \Vert_1$ is $\Vert \cdot \Vert_{\infty}$, and the last inequality holds since $\lh_t \in [0, 1]^K$.

\subsection{Proof of Lemma \ref{lm:w-t-t1}}
The proof is similar to that of Lemma \ref{lm:pf-thm1:w-t-t1} in Appendix \ref{app:lm:pf-thm1:w-t-t1}.
Fix $t \in [T]$. Focusing on Step $3$ of Algorithm \ref{alg:OCS} and following the same derivation as in (\ref{eq:model-der}), we have
\begin{equation*}
   \begin{split}
      \w_{t} & = \mathop{\arg\min}_{\w \in \widetilde \Delta_K} ~ \langle \w, \eta \lh_{t-1} \rangle + \D_{\phi}(\w \Vert \widetilde \w_{t}) \\
      & = \mathop{\arg\min}_{\w \in \widetilde \Delta_K} ~ \langle \w, \eta \lh_{t-1} \rangle + \phi(\w) - \langle \w, \nabla \phi(\widetilde \w_{t}) \rangle \\
      & = \mathop{\arg\max}_{\w \in \widetilde \Delta_K} ~ \langle \w, \nabla \phi(\widetilde \w_t) - \eta \lh_{t-1} \rangle - \phi(\w) \\
      & = \mathop{\arg\max}_{\w \in \widetilde \Delta_K} ~ \langle \w, \nabla \phi(\widetilde \w_t) - \eta \lh_{t-1} \rangle - \widehat \phi(\w) \\
      & = \nabla \widehat \phi^* (\nabla \phi(\widetilde \w_t) - \eta \lh_{t-1}) .
   \end{split}
\end{equation*}
Similarly, by Step $5$ of Algorithm \ref{alg:OCS}, we also have
\begin{equation*}
   \begin{split}
      \widetilde \w_{t+1} & = \mathop{\arg\max}_{\w \in \widetilde \Delta_K} ~ \langle \w, \nabla \phi(\widetilde \w_t) - \eta \lh_t \rangle - \widehat \phi(\w) \\
      & = \nabla \widehat \phi^* (\nabla \phi(\widetilde \w_t) - \eta \lh_t) .
   \end{split}
\end{equation*}
Utilizing the above two equalities and realizing that the dual norm of $\Vert \cdot \Vert_1$ is $\Vert \cdot \Vert_{\infty}$, we finish the proof as follows:
\begin{equation*}
   \begin{split}
      & ~ \langle \w_t - \widetilde \w_{t+1}, \lh_t - \lh_{t-1} \rangle \\ 
      = & ~ \langle \nabla \widehat \phi^* (\nabla \phi(\widetilde \w_t) - \eta \lh_{t-1}) - \nabla \widehat \phi^* (\nabla \phi(\widetilde \w_t) - \eta \lh_t), \lh_t - \lh_{t-1} \rangle \\
      \leq & ~ \Vert \nabla \widehat \phi^* (\nabla \phi(\widetilde \w_t) - \eta \lh_{t-1}) - \nabla \widehat \phi^* (\nabla \phi(\widetilde \w_t) - \eta \lh_t) \Vert_1 \Vert \lh_t - \lh_{t-1} \Vert_{\infty} \\
      \leq & ~ \Vert (\nabla \phi(\widetilde \w_t) - \eta \lh_{t-1}) - (\nabla \phi(\widetilde \w_t) - \eta \lh_t) \Vert_{\infty} \Vert \lh_t - \lh_{t-1} \Vert_{\infty} \\
      = & ~ \eta \Vert \lh_t - \lh_{t-1} \Vert_{\infty}^2
   \end{split}
\end{equation*}
where the first inequality is due to the Cauchy–Schwarz inequality, and the second inequality follows from (\ref{eq:prop2-conj}).

\subsection{Proof of Lemma \ref{lm:mtx-baru-u}}
Fix $t \in [T]$. By the definition of $\bar U_t^*$ in (\ref{eq:def:barut}), we have
\begin{equation}
   \label{eq:tmx-lm-baruu-c1}
   \begin{split}
      \tr\big((\bar U_t^* - U_t^*) Z_t\big) = \frac{ S \tr \big( ( -KU_t^* + I ) Z_t \big) }{TK} = \frac{ S \tr \big( -U_t^* Z_t \big) }{T} + \frac{ S \tr \big( Z_t \big) }{TK} .
   \end{split}
\end{equation}
Since $U_t^* \in \Omega_K$ is positive semidefinite, the eigenvalues of $U_t^*$ are all non-negative, which implies 
$$ \Vert U_t^* \Vert = \sum_{i=1}^K \vert \lambda_i(U_t^*) \vert = \sum_{i=1}^K \lambda_i(U_t^*) = \tr (U_t^*) = 1 $$
where $\lambda_i(\cdot)$ denotes the $i$-th eigenvalue.
Combining this with the fact that $\Vert Z_t \Vert_{*} \leq 1$ and $\Vert \cdot \Vert$ is the dual norm of $\Vert \cdot \Vert_{*}$, by the Cauchy–Schwarz inequality, we get
\begin{equation}
   \label{eq:tmx-lm-baruu-c2}
   \tr \big( -U_t^* Z_t \big) \leq \Vert -U_t^* \Vert \Vert Z_t \Vert_* = \Vert U_t^* \Vert \Vert Z_t \Vert_* \leq 1 .
\end{equation}
On the other hand, we have
\begin{equation}
   \label{eq:tmx-lm-baruu-c3}
   \tr(Z_t) = \sum_{i=1}^K \lambda_i(Z_t) \leq K .
\end{equation}
We finish the proof by combining (\ref{eq:tmx-lm-baruu-c1}), (\ref{eq:tmx-lm-baruu-c2}), and (\ref{eq:tmx-lm-baruu-c3}).

\subsection{Proof of Lemma \ref{lm:PCSP-1}}
Given a $K \times K$ symmetric and real matrix $W$, let $W=V \Lambda V^T$ be the eigendecomposition of $W$, where $V$ is an orthogonal matrix whose columns are the eigenvectors of $W$, and $\Lambda$ is a diagonal matrix whose entries are the eigenvalues of $W$. We define $\exp{(\Lambda)}$ to be a diagonal matrix with $\big(\exp{(\Lambda)}\big)_{ii} = \exp{(\Lambda_{ii})}$ and define $\exp{(W)}$ by
\begin{equation*}
   \exp{(W)} = V\exp{(\Lambda})V^T.
\end{equation*}
Following the proof of Theorem \ref{thm:PCS} in Appendix \ref{sec:pf-pcs}, we introduce $P_{t+1} \in \Omega_K$ defined by
\begin{equation*}
   P_{t+1} = \frac{\exp\big(\log{W_t} + \log{(I-\eta Z_t)}\big)}{\tr \Big( \exp\big(\log{W_t} + \log{(I-\eta Z_t)}\big) \Big)}
\end{equation*}
and rewrite $A_t$ as
\begin{equation}
   \label{eq:pf-mtx-split-at}
   \begin{split}
      A_t & = \tr\big( (W_t - W_{t+1}) (\eta Z_t) \big)  + \tr \big(W_{t+1} (\eta Z_t + \log{(I - \eta Z_t)} - \log{W_{t+1}} +  \log{W_t}) \big) \\
      & = \tr \big (\eta W_t Z_t \big) + \tr \big( W_{t+1}( \log{(I - \eta Z_t)} - \log{W_{t+1}} +  \log{W_t} ) \big) \\
      & = \tr \big (\eta W_t Z_t \big) + \tr \big( W_{t+1}( \log{(I - \eta Z_t)} - \log{P_{t+1}}  +  \log{W_t} ) \big) \\
      & ~~~~\, + \tr \big( W_{t+1} (\log{P_{t+1}} - \log{W_{t+1}} ) \big) \\
      & = \tr \big (\eta W_t Z_t \big) + \tr \big( W_{t+1}( \log{(I - \eta Z_t)} - \log{P_{t+1}}  +  \log{W_t} ) \big) - \D_{\psi} (W_{t+1} \Vert P_{t+1})
   \end{split}
\end{equation}
where the last equality follows from the definition of Bregman divergence with respect to $\psi$ in (\ref{def:breg-mtx}). 

Define $Q_t = \log{W_t} + \log{(I-\eta Z_t)}$. Let $Q_t = V \Lambda V^T$ be the eigendecomposition of $Q_t$. We have
\begin{equation}
   \label{eq:pt-1}
      P_{t+1} = \frac{\exp(Q_t)}{\tr \big( \exp(Q_t) \big)} = \frac{\exp( V \Lambda V^T )}{\tr \big(  \exp( V \Lambda V^T ) \big)} = \frac{V \exp(\Lambda) V^T}{ \tr \big(  V \exp{(\Lambda)} V^T \big) } .
\end{equation}
Since $\big(V \exp{(\Lambda)} V^T\big)V = \big(V \exp{(\Lambda)} \big) \big( V^T V \big) = V \exp{(\Lambda)}$, we know that the entries of the diagonal matrix $\exp{(\Lambda)}$ are the eigenvalues of $V \exp{(\Lambda)} V^T$ and thus
\begin{equation}
   \label{eq:equ-Lamda-T}
   \tr\big(V \exp{(\Lambda)} V^T \big) = \sum_{i=1}^K \lambda_i \big(V \exp{(\Lambda)} V^T \big) = \tr \big( \exp{(\Lambda)} \big)
\end{equation}
where recall that $\lambda_i(\cdot)$ denotes the $i$-th eigenvalue.
Substituting the above equality into (\ref{eq:pt-1}), we get
\begin{equation*}
   P_{t+1} = \frac{V \exp(\Lambda) V^T}{ \tr \big( \exp{(\Lambda)} \big) } = V \frac{ \exp(\Lambda) }{ \tr \big( \exp{(\Lambda)} \big) }  V^T .
\end{equation*}
Denoting $r = \tr \big( \exp{(\Lambda)} \big)$, we have
\begin{equation*}
   \begin{split}
      \log{P_{t+1}} & = V \log{\left( \frac{ \exp(\Lambda) }{ r } \right)}  V^T = V \big(\Lambda - (\log{ r } ) I\big) V^T \\
      & = V \Lambda V^T - (\log{ r }) V V^T = Q_t - (\log{ r }) I
   \end{split}
\end{equation*}
which, together with the definition of $Q_t$, implies
\begin{equation*}
   \log{(I - \eta Z_t)} + \log{W_t} - \log{P_{t+1}}  = Q_t - \log{P_{t+1}} = (\log{ r }) I
\end{equation*}
and hence
\begin{equation}
   \label{eq:pf-mtx-tr-wt-logrt}
   \tr \big( W_{t+1}( \log{(I - \eta Z_t)} - \log{P_{t+1}}  +  \log{W_t} ) \big) = (\log{ r }) \tr \big( W_{t+1} \big) = \log{ r }
\end{equation}
where the last equality holds since $W_{t+1}$ belongs to the clipped spectraplex $\widetilde \Omega_K$ defined in (\ref{def:clipped-spx}).

It remains to investigate the upper bound of $r$. To this end, by (\ref{eq:equ-Lamda-T}) and the definition of $Q_t$, we rewrite $r$ as
\begin{equation}
   \label{eq:pf-r-t}
   \begin{split}
      r = \tr\big(V \exp{(\Lambda)} V^T \big) = \tr \big( \exp{(Q_t)}  \big) = \tr \Big(  \exp{\big( \log{W_t} + \log{(I-\eta Z_t)} \big)} \Big) .
   \end{split}
\end{equation}
To proceed, we introduce the Golden-Thompson inequality \citep{golden1965lower,thompson1965inequality}: for any symmetric matrices $A$ and $B$,
\begin{equation*}
   \tr \big( \exp{(A+B)} \big) \leq \tr \big( \exp{(A)} \exp{(B)} \big) .
\end{equation*}
Applying this inequality to (\ref{eq:pf-r-t}) gives
\begin{equation}
   \label{eq:pf-mtx-r-t-up}
   r \leq \tr \Big( 
   \exp{\big( \log{W_t} \big)} \exp{\big( \log{(I - \eta Z_t)} \big)}  \Big) = \tr \Big( W_t (I - \eta Z_t) \Big) = 1 - \tr \big( \eta W_t Z_t \big)
\end{equation}
where the last equality holds since $\tr(W_t I) = \tr(W_t) = 1$.
Combining (\ref{eq:pf-mtx-r-t-up}) with (\ref{eq:pf-mtx-tr-wt-logrt}) and (\ref{eq:pf-mtx-split-at}), we get
\begin{equation*}
   \begin{split}
      A_t & = \tr ( \eta W_t Z_t ) + \log{r} - \D_{\psi} (W_{t+1} \Vert P_{t+1}) \\
      & \leq \tr ( \eta W_t Z_t ) + \log{\big( 1 - \tr ( \eta W_t Z_t ) \big)} - \D_{\psi} (W_{t+1} \Vert P_{t+1})\\
      & \leq \tr ( \eta W_t Z_t ) + \log{\big( 1 - \tr ( \eta W_t Z_t ) \big)}
   \end{split}
\end{equation*}
where the last inequality holds since Bregman divergence is always non-negative.

Finally, note that $\Vert W_t \Vert = 1, \Vert Z_t \Vert_{*} \leq 1$ and $\Vert \cdot \Vert$ is the dual norm of $\Vert \cdot \Vert_{*}$. Application of the Cauchy–Schwarz inequality gives
\begin{equation*}
   \tr(\eta W_t Z_t) = \eta \tr(W_t Z_t) \leq \eta \Vert W_t \Vert \Vert Z_t \Vert_{*} \leq \eta \leq \frac{1}{2} .
\end{equation*}
We conclude the proof by recalling the well-known inequality: $x + \log{(1-x)} \leq 0, ~\forall x < 1$.

\subsection{Proof of Lemma \ref{lm:PCSP-2}}
We start by rewriting $B_t$ as
\begin{equation}
   \label{eq:pf-mtx-split-bt}
   \begin{split}
      B_t & = \tr \Big(-\bar U_t^*\big(\eta Z_t + \log{(I - \eta Z_t)} \big) \Big) \\
      & = \tr \Big(- U_t^*\big(\eta Z_t + \log{(I - \eta Z_t)} \big) \Big) + \tr \Big( \big( U_t^* -\bar U_t^*\big) \big(\eta Z_t + \log{(I - \eta Z_t)} \big) \Big) .
   \end{split}
\end{equation}
We first focus on bounding the last term. By the definition of $\bar U_t^*$ in (\ref{eq:def:barut}), we have
\begin{equation}
   \label{eq:pf-mtx-ut-barut}
   \begin{split}
      & \tr \Big( \big( U_t^* -\bar U_t^*\big) \big(\eta Z_t + \log{(I - \eta Z_t)} \big) \Big) \\
      = \, & \frac{ S \tr \Big( \big( KU_t^* - I \big) \big(\eta Z_t + \log{(I - \eta Z_t)} \big) \Big) }{TK} \\
      = \, & \frac{ S \tr \Big( U_t^*  \big(\eta Z_t + \log{(I - \eta Z_t)} \big) \Big) }{T} + \frac{S \tr \Big( -\eta Z_t - \log{(I - \eta Z_t)} \Big) }{TK} .
   \end{split}
\end{equation}
Let $Z_t = V \Lambda V^T$ be the eigendecomposition of $Z_t$. We have
\begin{equation*}
   \begin{split}
      \eta Z_t + \log{(I - \eta Z_t)} & = \eta V \Lambda V^T + \log{(I - \eta V \Lambda V^T)} = \eta V \Lambda V^T + \log{(V V^T - \eta V \Lambda V^T)} \\
      & = \eta V \Lambda V^T + \log{\big(V (I - \eta \Lambda) V^T \big)} = \eta V \Lambda V^T + V \log{(I - \eta \Lambda)} V^T \\
      & = V \Big( \eta \Lambda + \log{(I - \eta \Lambda)} \Big) V^T .
   \end{split}
\end{equation*}
For any $i \in [K]$, let $a_i$ be the $i$-th diagonal entry of $\Lambda$. Since $\Vert Z_t \Vert_* \leq 1$, we have $ \vert a_i \vert \leq 1$ and $ \vert \eta a_i \vert \leq \eta \leq 1/2$. Utilizing the inequality $x + \log{(1-x)} \leq 0, ~\forall x < 1$, we get
\begin{equation*}
   \eta a_i + \log{(1 - \eta a_i)} \leq 0
\end{equation*}
which implies the eigenvalues of $\eta Z_t + \log{(I - \eta Z_t)}$ are all non-positive, and $\eta Z_t + \log{(I - \eta Z_t)}$ is hence negative semidefinite. Combining this with the fact that $U_t^*$ is positive semidefinite, we conclude that the eigenvalues of $U_t^*  \big(\eta Z_t + \log{(I - \eta Z_t)} \big)$ are all non-positive, which implies
\begin{equation}
   \label{eq:pf-mtx-ut-utb-1}
   \tr \Big( U_t^*  \big(\eta Z_t + \log{(I - \eta Z_t)} \big) \Big) = \sum_{i=1}^K \lambda_i \Big( U_t^*  \big(\eta Z_t + \log{(I - \eta Z_t)} \big) \Big) \leq 0 .
\end{equation}
On the other hand, by the inequality $-x - \log{(1-x)} \leq \vert x \vert / 2, ~\forall x \in [-1/2, 1/2]$, we have
\begin{equation*}
   -\eta a_i - \log{(1 - \eta a_i)} \leq \frac{\vert \eta a_i \vert}{2} \leq \frac{\eta}{2}
\end{equation*}
and
\begin{equation*}
   \begin{split}
      \tr \Big( -\eta Z_t - \log{(I - \eta Z_t)} \Big) & = \tr \Big( V \big( -\eta \Lambda - \log{(I - \eta \Lambda)} \big) V^T  \Big) \\
      & = \sum_{i=1}^K \big( -\eta a_i - \log{(1 - \eta a_i)} \big) \\
      & \leq  \sum_{i=1}^K \frac{ \eta }{2} = \frac{K\eta}{2} .
   \end{split}
\end{equation*}
Combining the above inequality with (\ref{eq:pf-mtx-ut-utb-1}) and (\ref{eq:pf-mtx-ut-barut}) gives
\begin{equation}
   \label{eq:pf-mtx-bt-2}
   \tr \Big( \big( U_t^* -\bar U_t^*\big) \big(\eta Z_t + \log{(I - \eta Z_t)} \big) \Big) \leq \frac{\eta S}{2T} .
\end{equation}
We now turn to bound $\tr \Big(- U_t^*\big(\eta Z_t + \log{(I - \eta Z_t)} \big) \Big)$.  To this end, we introduce the following fact \citep{steinhardt2014adaptivity}: for any symmetric and real matrix $X$ satisfying $-\frac{I}{2} \preceq X \preceq \frac{I}{2}$, we have
\begin{equation*}
   -X - X^2 \preceq \log{(I - X)}
\end{equation*}
where $A \preceq B$ means that $B - A$ is positive semidefinite. Since $\Vert Z_t \Vert_* \leq 1$ and $\eta \in (0, 1/2]$, we know that the maximum absolute eigenvalue of $\eta Z_t$ is not more than $1/2$, i.e., $\max_{i \in [K]} \vert \lambda_i (\eta Z_t) \vert \leq 1/2$. Therefore, we have $-\frac{I}{2} \preceq \eta Z_t \preceq \frac{I}{2}$ and
\begin{equation*}
   -\eta Z_t - \eta^2 Z_t^2 \preceq \log{(I - \eta Z_t)}
\end{equation*}
which implies $\log{(I - \eta Z_t)} + \eta Z_t + \eta^2 Z_t^2 $ is positive semidefinite. Combining this with the fact that $-U_t^*$ is negative semidefinite, we conclude that the eigenvalues of $- U_t^*\big( \log{(I - \eta Z_t)} + \eta Z_t + \eta^2 Z_t^2 \big)$ are all non-positive and hence
\begin{equation*}
   \tr \Big( - U_t^*\big( \log{(I - \eta Z_t)} + \eta Z_t + \eta^2 Z_t^2 \big) \Big) \leq 0.
\end{equation*}
Rearranging the above inequality, we obtain
\begin{equation*}
   \tr \Big( - U_t^*\big( \eta Z_t +  \log{(I - \eta Z_t)} \big) \Big) \leq \eta^2 \tr  \Big(U_t^* Z_t^2 \Big) .
\end{equation*}
Substituting the above inequality and (\ref{eq:pf-mtx-bt-2}) into (\ref{eq:pf-mtx-split-bt}) completes the proof.

\subsection{Proof of Lemma \ref{lm:logkt-mtx}}
We start by proving the following fact: for any $W \in \widetilde \Omega_K$, we have
\begin{equation}
   \label{pf-mt-fac}
   -\log{(KT/S)} \leq \lambda_i (\log{W}) \leq 0, ~\forall i \in [K] .
\end{equation}
where $\lambda_i(\cdot)$ denotes the $i$-th eigenvalue.

\BeginProof Fix $W \in \widetilde \Omega_K$. Let $W = V\Lambda V^T$ be the eigendecomposition of $W$. It follows that
\begin{equation*}
   \log{W} = V (\log{\Lambda}) V^T
\end{equation*}
which implies that the diagonal entries of $\log{\Lambda}$ are the eigenvalues of $\log{W}$. For any $i \in [K]$, let $a_i = \Lambda_{ii}$ denote the $i$-th diagonal entry of $\Lambda$. Since $a_i$ is the eigenvalue of $W \in \widetilde \Omega_K$, by the definition of $\widetilde \Omega_K$ in (\ref{def:clipped-spx}), we have
\begin{equation*}
   \frac{S}{TK} \leq a_i \leq 1
\end{equation*}
and hence
\begin{equation*}
   -\log{(KT/S)} \leq \log{a_i} \leq 0.
\end{equation*}
We finish the proof by noticing that $\lambda_i(\log{W}) = (\log{\Lambda})_{ii} = \log{(\Lambda_{ii})} = \log{a_i} $. \EndProof

We are now ready to prove Lemma \ref{lm:logkt-mtx}. Fix $X, Y, Z \in \widetilde \Omega_K$. First, applying (\ref{pf-mt-fac}) to $Y$ and $Z$ indicates that
$\log{Y}$ is negative semidefinite and $\log{Z}$ satisfies
\begin{equation*}
   \Vert \log{Z} \Vert_* = \max_{i \in [K]} \vert \lambda_i (\log{Z}) \vert \leq \log{(KT/S)} .
\end{equation*}
Then, we expand $\tr \big(X (\log{Y} - \log{Z}) \big)$ as
\begin{equation}
   \label{pf-logktm-c1}
   \tr \big(X (\log{Y} - \log{Z}) \big) = \tr \big(X \log{Y}  \big) + \tr \big(-X \log{Z} \big) .
\end{equation}
Since $X \in \widetilde \Omega_K$ is positive semidefinite, we conclude that the eigenvalues of $X \log{Y}$ are all non-positive and thus
\begin{equation}
   \label{pf-logktm-c2}
   \tr (X \log{Y}) = \sum_{i=1}^K \lambda_i (X \log{Y}) \leq 0.
\end{equation}
Finally, application of the Cauchy–Schwarz inequality gives
\begin{equation}
   \label{pf-logktm-c3}
   \tr (-X \log{Z}) \leq \Vert - X \Vert \Vert \log{Z} \Vert_{*} = \Vert X \Vert \Vert \log{Z} \Vert_* = \Vert \log{Z} \Vert_* \leq \log{(KT/S)} . 
\end{equation}
Combining (\ref{pf-logktm-c1}), (\ref{pf-logktm-c2}), and (\ref{pf-logktm-c3}) completes the proof.

\end{document}